\RequirePackage{fix-cm}
\documentclass[twocolumn]{svjour3}          

\usepackage{amsmath}
\usepackage{adjustbox}

\newcommand{\eg}{\textit{e}.\textit{g}.}
\newcommand{\aka}{\textit{a}.\textit{k}.\textit{a}.}

\usepackage{times}

\usepackage{hyperref}
\hypersetup{
colorlinks=true,
linkcolor=red,
citecolor=green
}
\usepackage{url}

\usepackage{amsmath}
\usepackage{amssymb}
\usepackage{arydshln}
\usepackage{booktabs} 
\usepackage{multirow}
\usepackage{multicol}
\usepackage{color, colortbl}
\usepackage{adjustbox}
\usepackage{xspace}
\newcolumntype{M}[1]{>{\centering\arraybackslash}m{#1}}

\usepackage{arydshln}
\usepackage{pifont}
\newcommand{\xmark}{\ding{55}}

\usepackage[numbers]{natbib}

\smartqed  
\begin{document}
\sloppy
\title{Multi-Modal Few-Shot Object Detection with Meta-Learning-Based Cross-Modal Prompting}


\author{Guangxing Han, Long Chen, Jiawei Ma, Shiyuan Huang, \\ Rama Chellappa, Shih-Fu Chang}


\institute{Guangxing Han \at
              Columbia University \\
              \email{gh2561@columbia.edu}
        \and
        Long Chen \at
              Columbia University \\
              \email{zjuchenlong@gmail.com}
        \and
        Jiawei Ma \at
              Columbia University \\
              \email{jiawei.m@columbia.edu}
        \and
        Shiyuan Huang \at
              Columbia University \\
              \email{sh3813@columbia.edu}  
        \and
        Rama Chellappa \at
              Johns Hopkins University \\
              \email{rchella4@jhu.edu}
        \and 
        Shih-Fu Chang \at
              Columbia University \\
              \email{sc250@columbia.edu}
}

\date{Received: date / Accepted: date}

\maketitle
\begin{abstract}
We study multi-modal few-shot object detection (FSOD) in this paper, using both few-shot visual examples and class semantic information for detection, which are complementary to each other by  definition. Most of the previous works on multi-modal FSOD are fine-tuning-based which are inefficient for online applications. Moreover, these methods usually require expertise like class names to extract class semantic embedding, which are hard to get for rare classes. Our approach is motivated by the high-level conceptual similarity of (metric-based) meta-learning and prompt-based learning to learn generalizable few-shot and zero-shot object detection models respectively without fine-tuning. Specifically, we combine the few-shot visual classifier and text classifier learned via meta-learning and prompt-based learning respectively to build the multi-modal classifier and detection models. In addition, to fully exploit the pre-trained language models, we propose meta-learning-based cross-modal prompting to generate soft prompts for novel classes present in few-shot visual examples, which are then used to learn the text classifier. Knowledge distillation is introduced to learn the soft prompt generator without using human prior knowledge of class names, which may not be available for rare classes. Our insight is that the few-shot support images naturally include related context information and semantics of the class. We comprehensively evaluate the proposed multi-modal FSOD models on multiple few-shot object detection benchmarks, achieving promising results.
\end{abstract}

\section{Introduction}
\label{sec:intro}

\begin{figure*}[t]
\begin{center}
\includegraphics[scale=0.61]{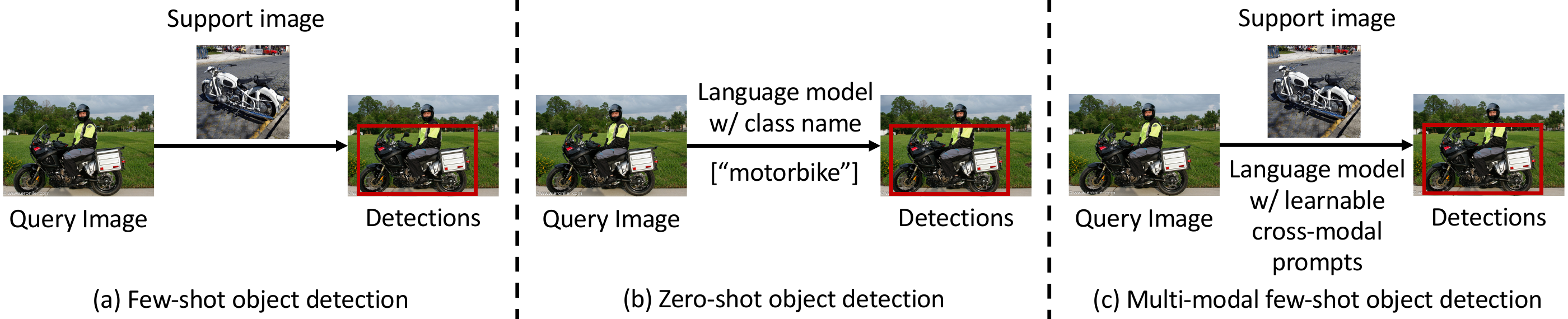} 
\end{center}
\caption{Comparisons of zero-shot object detection, few-shot object detection, and our multi-modal few-shot object detection.}
\label{figure_1}
\end{figure*}

Object detection is one of the most fundamental tasks in computer vision. Recently, deep learning-based methods~\cite{ren2015faster,redmon2016you,liu2016ssd,carion2020end} have achieved great progress in this field. However, these methods usually need to collect large-scale labeled training data with bounding box annotations for each class, which is time-consuming and expensive, especially for rare classes. In order to reduce the number of labeled training data needed for learning, \textbf{few-shot} learning-based methods~\cite{wang2020few,wu2020multi,Sun_2021_CVPR,Zhu_2021_CVPR,kang2019few,fan2020few,yan2019meta,han2021meta,Han_2021_ICCV,Han_2022_CVPR} and \textbf{zero-shot} learning-based methods~\cite{bansal2018zero,zareian2021open,joseph2021towards,gu2022openvocabulary} are proposed to detect novel categories using few-shot visual examples and class semantic information (\eg, attributes or word embeddings), respectively. 

Few-shot object detection (\textbf{FSOD}) methods~\cite{wang2020few,wu2020multi,Sun_2021_CVPR,Zhu_2021_CVPR,kang2019few,fan2020few,yan2019meta,han2021meta,Han_2021_ICCV,Han_2022_CVPR} are developed to detect objects using only a few visual training examples. Meta-learning-based FSOD methods~\cite{fan2020few,yan2019meta,han2021meta,Han_2021_ICCV,Han_2022_CVPR} have been shown to be effective for learning class-agnostic metric-space over data-abundant base classes, which can be generalized to few-shot novel classes without fine-tuning, and have been widely used for FSOD. 

On the other hand, zero-shot object detection (\textbf{ZSD}) methods~\cite{bansal2018zero,zareian2021open,joseph2021towards,gu2022openvocabulary} usually leverage auxiliary class semantic information (\eg, attributes or word embeddings) to detect unseen object categories which do not have any visual training samples, by aligning the visual-semantic feature space during training. Recently, large-scale vision-language pre-training~\cite{Su2020VL-BERT:,vilt,tan2019lxmert,radford2021learning} has demonstrated a strong ability to learn aligned cross-modal representations. Existing ZSD works~\cite{radford2021learning,gu2022openvocabulary,li2022languagedriven} propose to recognize unseen object categories by exploiting the aligned visual-semantic feature space from pre-trained vision-language models (\eg, CLIP~\cite{radford2021learning}) and constructing category text classifiers via prompt-based learning~\cite{liu2021pre}.

However, most of the previous works focus on learning under either a few-shot or zero-shot setting. In fact, the visual and semantic feature spaces have different structures by definition and could be complementary to each other~\cite{xing2019adaptive}. As shown in Fig.~\ref{figure_1}(a), the few-shot visual examples, containing more local and fine-grained details, share the same embedding space with the query image. Meanwhile, as illustrated in Fig.~\ref{figure_1}(b), class semantic information offers high-level abstraction and could have better generalization ability compared with few-shot visual examples. Multi-modal FSOD, in Fig.~\ref{figure_1}(c), aims to leverage both few-shot visual examples and class semantic information for detection. 

\begin{table}[t]
    \caption{Comparison with a previous multi-modal FSOD work~\cite{Zhu_2021_CVPR}. (1) As shown in the upper part of the table, both methods use few-shot visual data and the pre-trained language model for multi-modal FSOD. (2) The key difference is that our method is (metric-based) meta-learning-based and does not need fine-tuning, while SRR-FSD~\cite{Zhu_2021_CVPR} is fine-tuning-based. In addition, we do not need human prior knowledge of class names for novel classes, which may be rare and need expertise. Instead, we generate soft cross-modal prompts for novel classes based on few-shot support images, to extract the class semantic embedding. \textbf{S}: Similarities. \textbf{D}: Differences.}
    \addtolength{\tabcolsep}{-0.1cm}
    \adjustbox{width=\linewidth}{
    \begin{tabular}{c|c|c|c}
    \toprule
    \multicolumn{2}{c|}{} & SRR-FSD~\cite{Zhu_2021_CVPR}  & {Our method} \\ \midrule
    \multirow{2}{*}{\textbf{S}} & Visual data & Few-shot & Few-shot \\
    & Language model & \checkmark & \checkmark \\ \cdashline{1-4}
    \multirow{4}{*}{\textbf{D}} & Class name & \checkmark & \xmark \\
    & Text prompt & Fixed class name & Cross-modal prompt \\
    & Classifier & Text classifier & Multi-modal classifier \\
    & Add novel class & Fine-tuning & Meta-testing \\
    \bottomrule
    \end{tabular}}
    \label{tab:compare_srr_fsd}
\end{table}

\begin{figure*}[t]
\begin{center}
\includegraphics[scale=0.72]{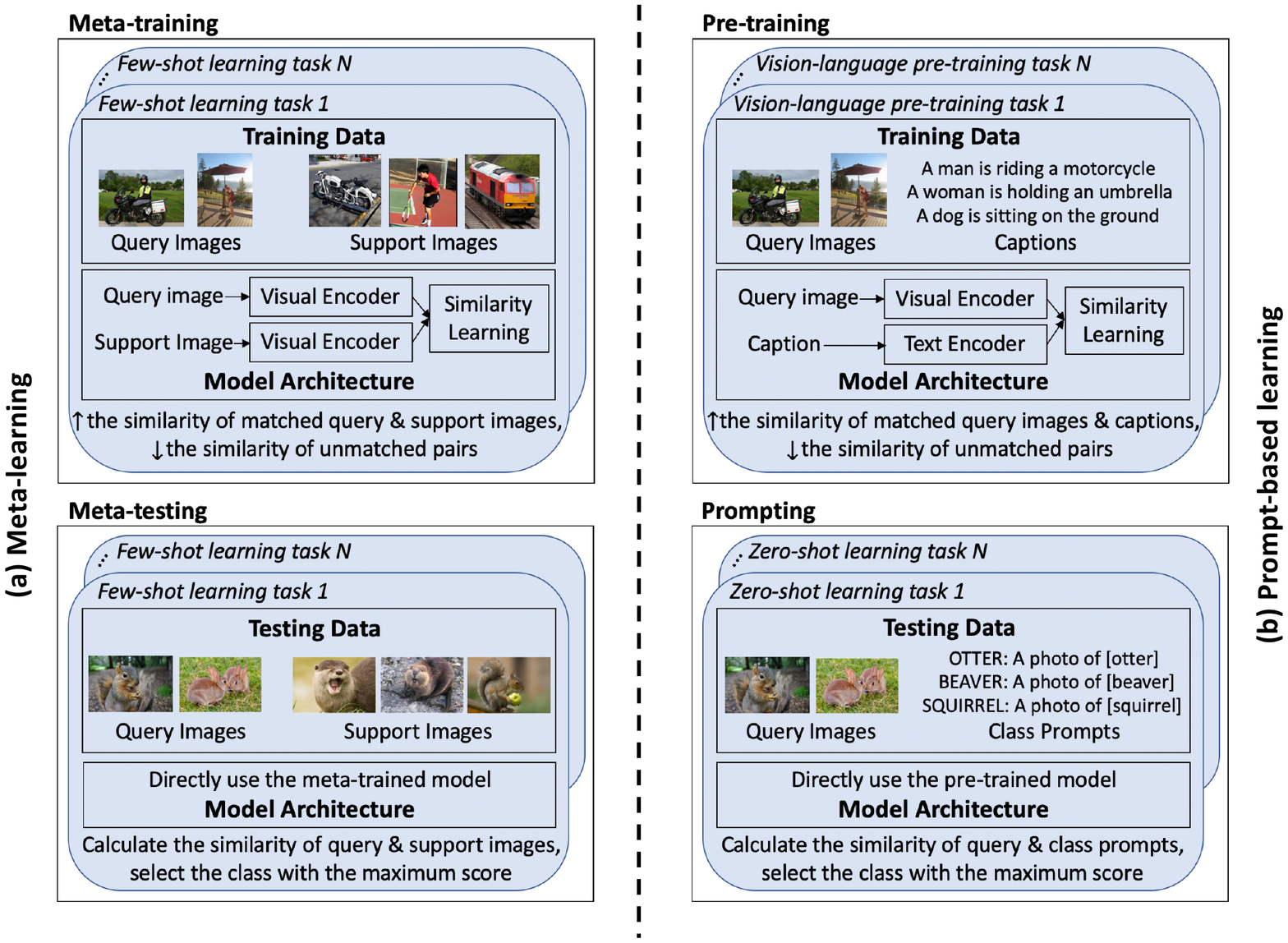} 
\end{center}
\caption{Comparison of (metric-based) meta-learning and prompt-based learning. The two learning paradigms both have consistent task formulations and model architectures for training and testing, such that they can reuse the trained models for new classes and tasks without fine-tuning. Motivated by this high-level conceptual similarity of two learning paradigms to learn generalizable few-shot and zero-shot learning models respectively, we propose to combine them for multi-modal FSOD without fine-tuning.}
\label{figure_2}
\end{figure*}

There are very few works on developing multi-modal FSOD. As shown in Table~\ref{tab:compare_srr_fsd}, one closely related work SRR-FSD~\cite{Zhu_2021_CVPR} is a fine-tuning-based method. It uses the class semantic embedding as the classifier, and trains the detector to project the objects from the visual space to the semantic space using few-shot visual training data. Despite large performance gains, there are two main weaknesses in this method. First, it needs additional model training to enroll novel classes to the system, which is inefficient and usually requires large computational resources. What's worse, it has the risk of overfitting under extremely few-shot scenarios, \eg, 1-shot. Second, it requires the class name of novel classes to extract the class semantic embedding. However, in the real world applications, annotators probably do not know the class names of the object they want to detect, which may be rare and needs expertise, while taking a few pictures of the object is much easier. Therefore, it is highly needed to develop generalizable multi-modal FSOD models without fine-tuning, and do not need strong human prior knowledge like class names.

\begin{figure}[t]
\begin{center}
\includegraphics[scale=0.33]{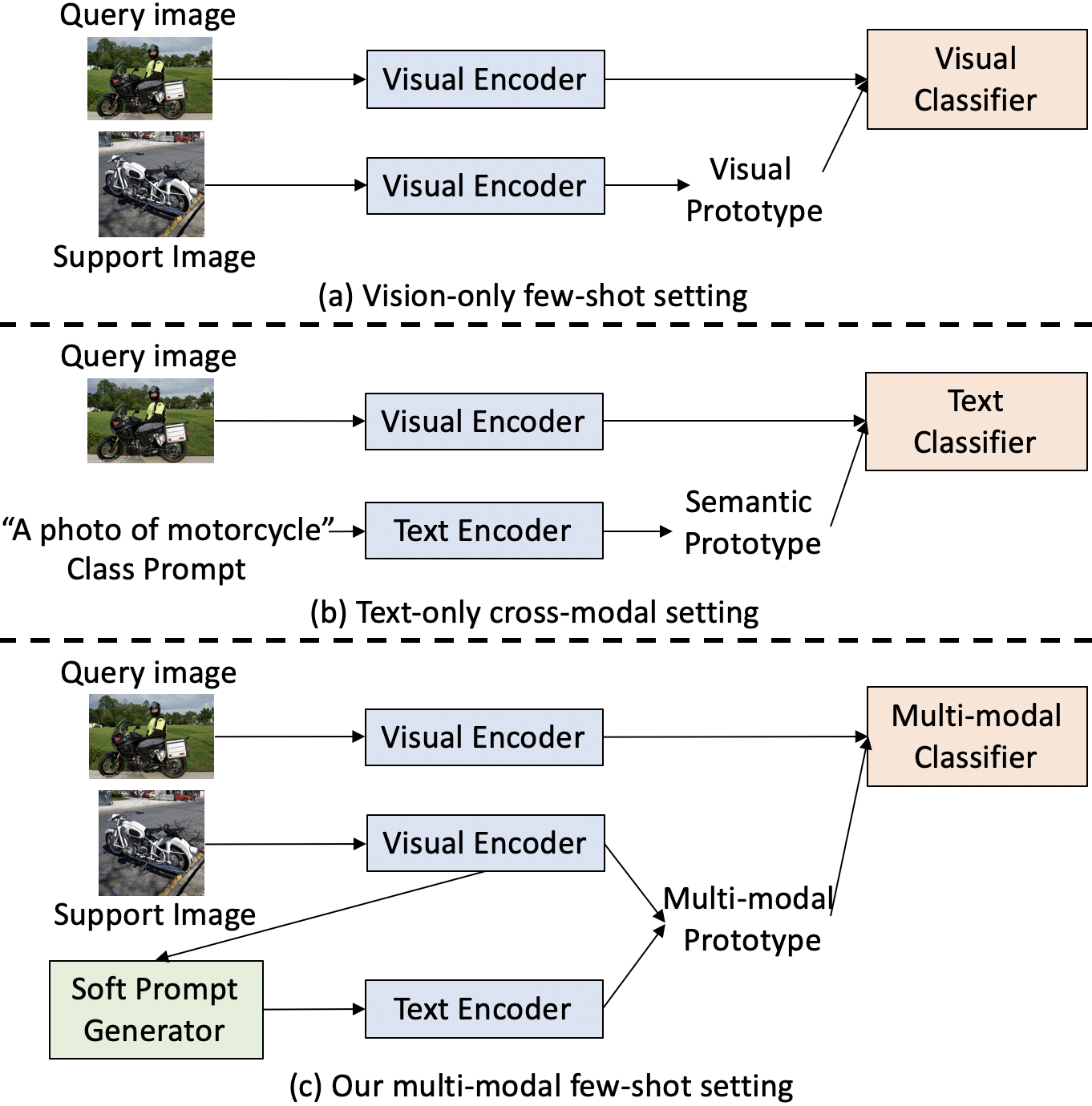} 
\end{center}
\caption{The key module of similarity learning in the vision-only few-shot, text-only cross-modal, and our proposed multi-modal few-shot settings. The visual/text/multi-modal classifiers are all metric-learning-based classifiers. By combining the learned few-shot visual and semantic prototype in meta-learning and prompt-based learning, we obtain the generalizable multi-modal prototype and classifier without fine-tuning. In addition, the proposed cross-model prompting module can generate soft prompt tokens for novel classes, based on the few-shot support images, without using any human prior knowledge.}
\label{figure_3}
\end{figure}

As shown in Fig.~\ref{figure_2}, our approach is inspired by the high-level conceptual similarity of the two learning paradigms meta-learning\footnote{We mainly study the metric-learning-based meta-learning methods for FSOD, which can be easily generalized to novel classes during meta-testing, without fine-tuning.} 
and prompting-based learning to learn generalizable few-shot and zero-shot object detection models without fine-tuning. 
Specifically, in meta-learning, both meta-training/-testing consist of multiple FSOD tasks (\aka, episodes). The metric-based meta-learning methods~\cite{fan2020few}, aim to learn class-agnostic few-shot visual classifier (prototype-based comparison network~\cite{snell2017prototypical,sung2018learning}) and FSOD models during meta-training, which can be generalized to novel classes during meta-testing without fine-tuning. In prompt-based learning (\eg, CLIP~\cite{radford2021learning}), zero-shot learning is reformulated as the image-text matching task, which is same as the pre-training task, and the pre-trained language model can be used to build text classifiers (\aka, class embedding) using the class prompts without fine-tuning.
As shown in Fig.~\ref{figure_3}, by combining the above learned few-shot visual and text classifiers, we can obtain the generalizable multi-modal classifier without fine-tuning. Compared with few-shot classification, FSOD is more challenging to handle both localization and classification tasks. Thus, we learn two multi-modal classifiers to generate class-specific proposals based on RPN~\cite{ren2015faster} and classify the proposals based on R-CNN~\cite{girshick2015fast}, respectively.

\begin{figure}[t]
\begin{center}
\includegraphics[scale=0.26]{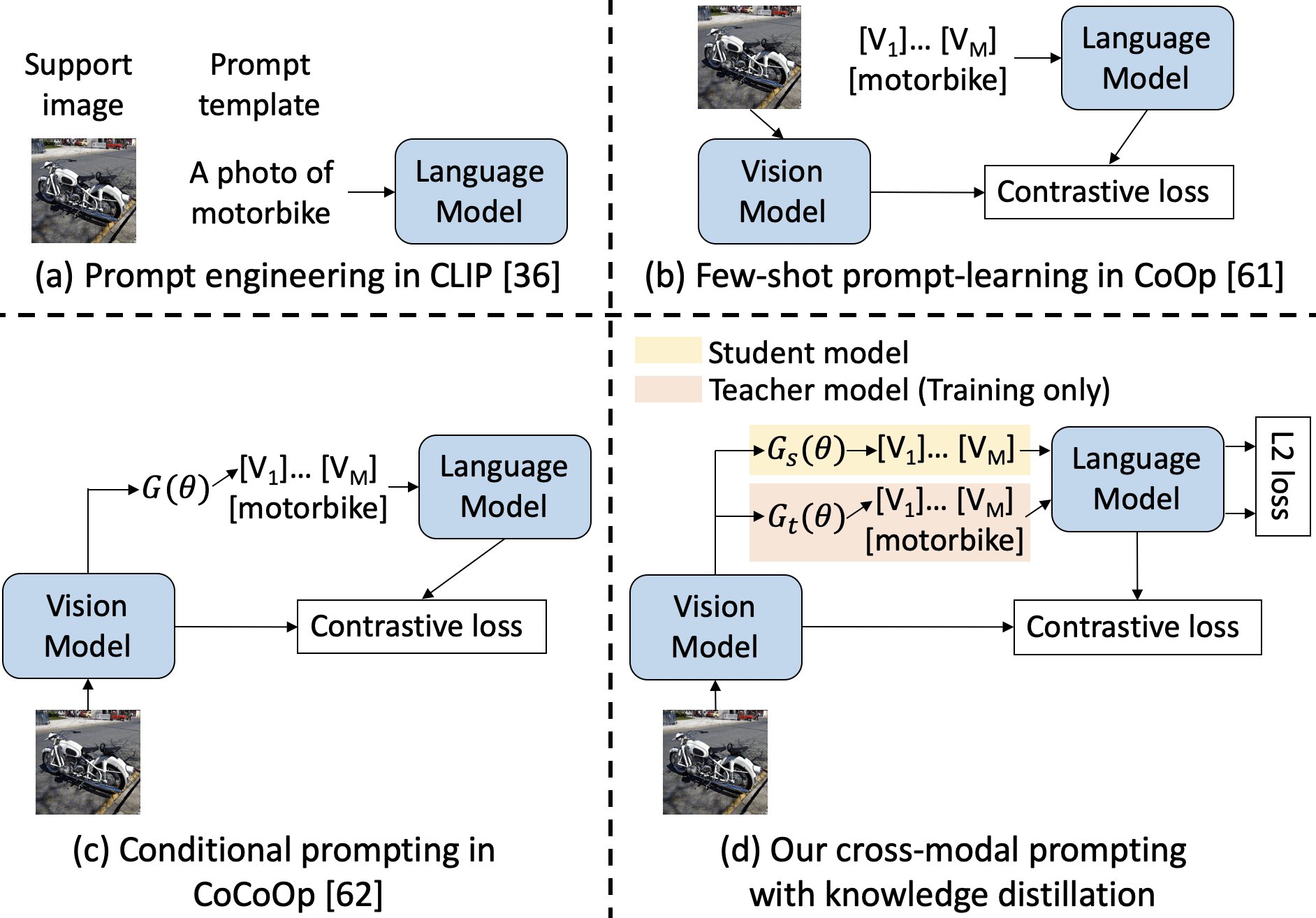} 
\end{center}
\caption{Comparisons of different prompting methods. $\{V_i\}_{i=1}^{M}$ are the learnable prompt tokens. $G(\theta)$ is the prompt generation module.}
\label{figure_4}
\end{figure}

How to design prompt tokens is crucial to fully exploit the pre-trained language model. 
As shown in Fig.~\ref{figure_4}, the previous work such as CLIP~\cite{radford2021learning} manually designs the prompt templates which requires expertise. CoOp~\cite{zhou2021learning} and CoCoOp~\cite{zhou2022conditional} in Fig.~\ref{figure_4}(b) and \ref{figure_4}(c) automatically learn the prompt tokens (\aka, soft prompts) using few-shot training examples. However, all these methods require the class names of novel classes which are combined with the soft prompt as the final prompt to the language model. In fact, it is hard to know the class names for rare classes.
To address this problem, we propose to generate contextual soft prompts for novel classes without class names (student model), by meta-training the soft prompt generator over base classes, conditioned on the few-shot support images. Our insight is that few-shot support images include useful context information and semantic knowledge of the class. Meanwhile, we also learn a teacher model using base-classes training data, which combines the class names with the generated soft prompt as the final prompt to the language model. Inspired by knowledge distillation~\cite{hinton2015distilling}, our student model is trained to mimic the class semantic embedding extracted by the teacher model. After meta-training, our student model can achieve competitive performance in novel classes compared with the teacher model.

Our contributions can be summarized as:
\begin{enumerate}
   \item 
   We combine the two learning paradigms, meta-learning and prompt-based learning to learn generalizable multi-modal FSOD models without fine-tuning and without using human prior knowledge of class names.
   \item 
   The meta-learning-based cross-modal prompting can generate soft prompts for novel classes using the few-shot visual examples. We also introduce knowledge distillation during training to learn the prompt generator, without using human prior knowledge like class names.
   \item 
   We evaluate the proposed model, denoted as \textbf{MM-FSOD}, on two widely used FSOD benchmarks (PASCAL VOC \& MSCOCO) and achieve promising results.
\end{enumerate}

\section{Related Work}
\label{related_work}

We first review the recent progress on object detection (including fully-supervised, few-shot, and zero-shot models), which is the major task in our work. Then we review meta-learning and prompt-based learning (including vision-language pre-training models and CLIP), which are closely related to our proposed models.

\subsection{Few-Shot and Zero-Shot Object Detection}

Despite the tremendous progress in object detection models, they usually need a sufficient amount of human annotations per class for model training, which is time-consuming and expensive. On the other hand, learning accurate object detection models with few training data, \eg, few-shot object detection and zero-shot object detection have attracted great interest from the community.

\textit{\textbf{Few-shot object detection}} aims to detect novel objects in the images using a few training examples (\aka, support images), with the help of data-abundant base classes. Existing works can be mainly grouped into the following two categories with different learning strategies: 

(1) Fine-tuning-based methods~\cite{wang2020few,wu2020multi,Sun_2021_CVPR,Zhang_2021_CVPR,Zhu_2021_CVPR}. They first train object detectors over base classes and then fine-tune the pre-trained detection models over few-shot novel classes, and usually utilize training strategies like re-sampling~\cite{wang2020few} and re-weighting~\cite{lin2017focal} to train models with the unbalanced training set of many-shot base-classes dataset and few-shot novel-classes dataset. 

(2) Meta-learning-based methods~\cite{kang2019few,fan2020few,yan2019meta,han2021meta,Han_2021_ICCV,Han_2022_CVPR,hsieh2019one}. 
Meta-learners~\cite{kang2019few,fan2020few,yan2019meta,han2021meta,Han_2022_CVPR,Han_2021_ICCV,hsieh2019one} are introduced to extract the meta knowledge over base classes which can be generalized to novel classes. 
Among them, metric-learning-based methods have been demonstrated to be effective for FSOD by learning a class-agnostic metric-space over base classes. To be specific, these methods employ a siamese network architecture and calculate the similarity between the query image regions and few-shot support images using metric-learning~\cite{karlinsky2019repmet}. Subsequent works propose multiple feature fusion networks~\cite{fan2020few,yan2019meta,xiao2020few}, feature alignment~\cite{han2021meta}, GCN~\cite{Han_2021_ICCV}, and non-local attention/transformer~\cite{wang2018non,hsieh2019one,chen2021dual,NEURIPS2020_fa28c6cd,chen2021adaptive,Han_2022_CVPR}) to improve the similarity learning between the query and few-shot support images. 

Metric-learning-based methods usually have stronger generalization ability compared to fine-tuning-based methods. The reason is that they do not learn a separate classifier for each of the classes (including base \& novel classes). Instead, they learn a shared class-agnostic metric-based classifier for all the classes.

\textit{\textbf{Zero-shot object detection}} (\aka, open-vocabulary object detection), learns to detect object categories that are not seen during training. Existing methods~\cite{bansal2018zero,zareian2021open,joseph2021towards,gu2022openvocabulary} solve this problem by first learning common visual-semantic feature space by large-scale vision-language pre-training~\cite{Su2020VL-BERT:,vilt,tan2019lxmert,radford2021learning}, and then learning the object detection models over seen classes based on the pre-trained aligned visual-semantic space. After that, the detection models can be applied to unseen classes using the class semantic features. OVR-CNN~\cite{zareian2021open} uses external image-caption pairs to learn a common visual-semantic space. ViLD~\cite{gu2022openvocabulary} distills the knowledge from a pre-trained open-vocabulary image classification model CLIP~\cite{radford2021learning} (teacher) into a two-stage detection model Faster R-CNN (student).

\subsection{Few-Shot Learning and Meta-Learning}
Few-shot learning aims to recognize novel classes using only a few examples. 
Meta-learning (\aka, learning to learn) has been shown to be a promising learning paradigm for few-shot learning tasks by transferring meta-knowledge learned from data-abundant base classes to data-scarce novel classes. Current meta-learning-based few-shot learning methods can be roughly divided into the following three categories according to the learned meta-knowledge: 

(1) Optimization-based methods~\cite{finn2017model,ma2022fewgaze}. These methods learn the optimization strategy as meta-knowledge. For example, Model-Agnostic Meta-Learning (MAML~\cite{finn2017model}) learns a good initialization so that the learner could rapidly adapt to novel tasks within a few optimization steps. 

(2) Parameter generation-based methods~\cite{gidaris2018dynamic,Huang_2022_CVPR}. These methods learn the parameter generation network as meta-knowledge. For example, \textit{Gidaris et al.}~\cite{gidaris2018dynamic} proposes to learn an attention-based weight generator to generate the classifier weights for novel classes. 

(3) Metric-learning-based methods~\cite{vinyals2016matching,snell2017prototypical,sung2018learning,Ma_2021_ICCV,ypsilantis2021met}. These methods learn a generalizable similarity metric-space as meta-knowledge. For example, Matching Networks~\cite{vinyals2016matching} can be interpreted as a weighted nearest-neighbor classifier with an attention mechanism over the learned embedding of the support images.
Prototypical Networks~\cite{snell2017prototypical} calculate the prototype of novel classes by averaging the features of a few samples, and then perform classification by nearest neighbor search. Relation Networks~\cite{sung2018learning} learn a distance metric network to calculate the similarity between the query image and few-shot class images. The metric-learning-based methods have been widely used for FSOD.

\subsection{Prompt-Based Learning}

Prompting-based learning~\cite{liu2021pre} has been proposed in the NLP community as an alternative solution to fine-tuning. GPT-3~\cite{GPT-3} first shows that language models pre-trained on large-scale datasets are few-shot learners without fine-tuning by reformulating the downstream tasks as masked language modeling tasks in pre-training (\aka, prompting), which can also reduce the objective gap between pre-training and downstream tasks. Since then, following the ``pre-train, prompt, and predict'' paradigm, various prompt design approaches are proposed, including hard prompt (discrete language phrases) and soft prompt (continuous learnable embeddings). Some works~\cite{schick-schutze-2021-exploiting,shin-etal-2020-autoprompt} focus on prompt engineering by automatically generating proper discrete prompts for downstream tasks. However, restricting prompts to discrete language tokens is usually sub-optimal. Prompt-tuning~\cite{lester-etal-2021-power,li-liang-2021-prefix} is proposed to replace the human-defined hard prompts with soft learnable prompt tokens. The soft prompts are learned for downstream tasks through back-propagation while freezing the pre-trained language model.

Similar to the “prompt engineering” in the NLP community, the performance of transferring pre-trained language models (\eg, BERT~\cite{devlin-etal-2019-bert}) or vision-language models (\eg, CLIP~\cite{radford2021learning}) to downstream vision-language tasks can be significantly improved by customizing the prompt text to each task~\cite{radford2021learning,zhou2021learning,yao2021cpt}. For example, \textit{Frozen}~\cite{tsimpoukelli2021multimodal} trains a vision encoder to represent each image as a sequence of continuous embeddings, such that a pre-trained, frozen language model prompted with this prefix generates the appropriate caption, and shows promising results in multiple vision-language downstream tasks. CLIP~\cite{radford2021learning} proposes prompt engineering and ensembling, which brings large improvement for zero-shot classification. CoOp~\cite{zhou2021learning} and CoCoOp~\cite{zhou2022conditional} apply the idea of continuous prompt learning to the vision-language pre-trained model CLIP, and shows improvements for few-shot classification. CPT~\cite{yao2021cpt} reformulates visual grounding into a fill-in-the-blank problem with color-based co-referential markers in image and text, and the bottleneck of this method is the limited number of the color set. 

Our method is closely related to the previous work \textit{Frozen}~\cite{tsimpoukelli2021multimodal}. \textit{Frozen} proposes to convert support images, text descriptions and query image into a sequence of tokens, which are fed into the pretrained language model for multi-modal few-shot classification. \textit{Frozen} can bind visual images with concepts implicitly by the language model, but lacks explainability. In contrast, our method for multi-modal FSOD has better explainability, by combining meta-learning and prompting at the classifier level, and constructing a multi-modal classifier for each class. Besides, the meta-learning-based cross-modal prompting at token level improves the text classifier with better prompts.

\section{The Proposed Approach}

\subsection{Task Definition}

Multi-modal few-shot object detection is built upon traditional vision-only FSOD~\cite{kang2019few,wang2020few}), and further introduce additional class semantic information to assist in detection for few-shot novel classes. 

Specifically, we have two sets of classes $C = C_{base} \cup C_{novel}$ and $C_{base} \cap C_{novel} = \emptyset$, where base classes $C_{base}$ have plenty of visual training examples per class, and novel classes $C_{novel}$ (\aka, support classes) only have very few visual training examples per class (\aka, support images). For $K$-shot (\eg, $K=1,5,10$) object detection, we have exactly $K$ bounding box annotations for each novel class $c \in C_{novel}$ as the training data.
Meanwhile, we also assume that we only know the class names for common many-shot base classes, but do not know the class names for few-shot novel classes because they are rare. 
We have the pre-trained language models to extract the class semantic features. 

The goal of multi-modal FSOD is to leverage the few-shot visual examples and the pre-trained language model to detect novel classes, with the assistance of data-abundant base-classes training data.

\begin{figure}[t]
\begin{center}
\includegraphics[scale=0.7]{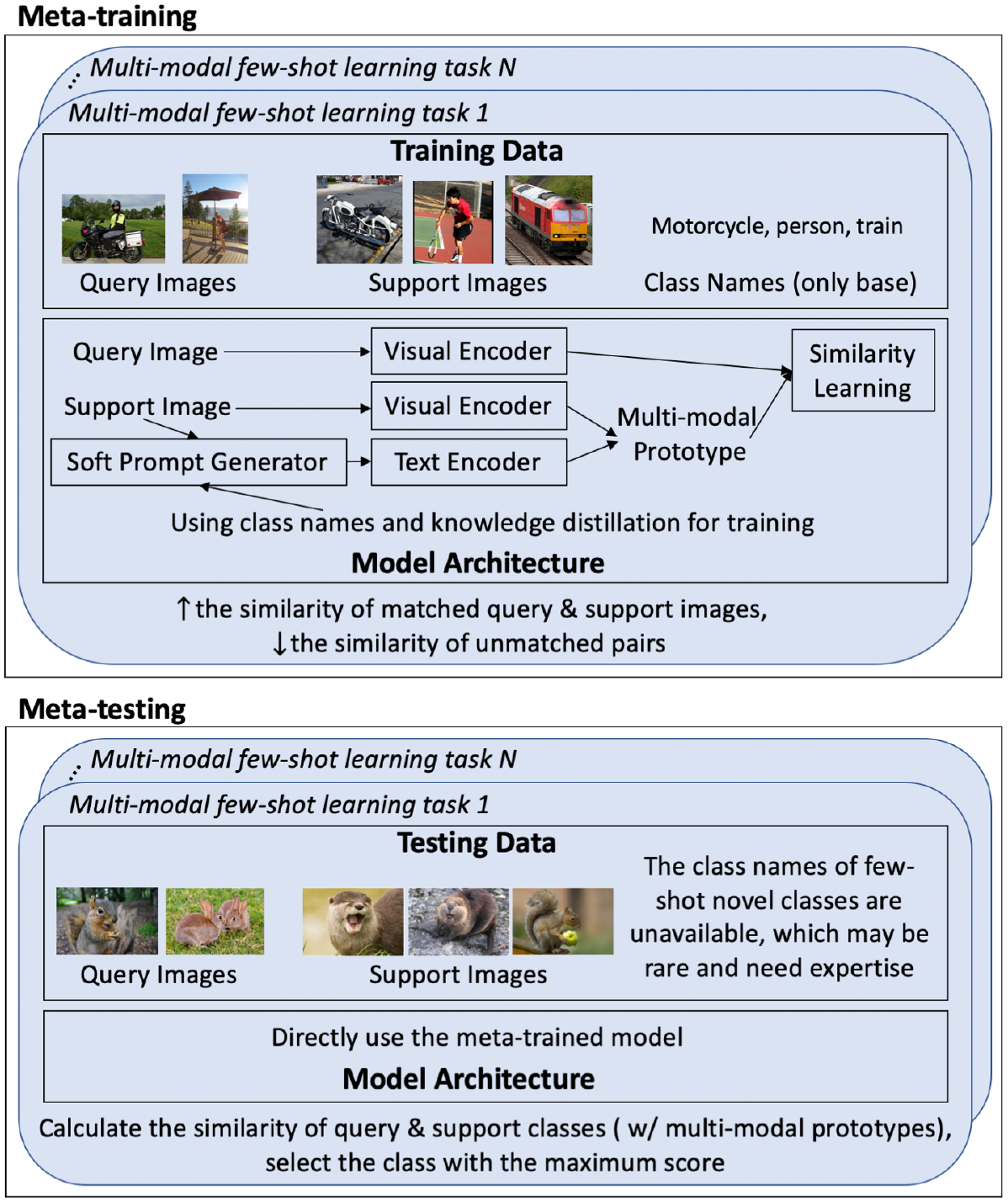} 
\end{center}
\caption{The overall idea of our proposed method. Our method is built upon the (metric-based) meta-learning framework. Our contributions are: (1) By combining the learned few-shot visual and text classifiers, we can obtain the generalizable multi-modal classifier for novel classes without fine-tuning. (2) To further reduce the dependency to human prior knowledge of class names, which are hard to get for rare classes, we propose the cross-model prompting to generate soft prompt tokens for novel classes and train the module using base-class dataset with class names and knowledge distillation.}
\label{figure_5}
\end{figure}

\subsection{Overview of Our Approach}

The goal of our work is to develop multi-modal FSOD models without fine-tuning, by learning transferable class-agnostic multi-modal FSOD models over many-shot base classes. Formally, as shown in Fig.~\ref{figure_5}, we sample multiple training episodes from the base class training data and learn our model via episode-based training, following previous works~\cite{fan2020few,han2021meta}. 
In each episode $\mathcal{D}=\{\mathcal{S}, \mathcal{Q}\}$, we have a $N$-way $K$-shot support set $\mathcal{S}$ and a query set $\mathcal{Q}$. The query set $\mathcal{Q}$ has the ground-truth bounding boxes for each of the $N$ categories. Meanwhile, we also have the class names $\{t_i\}^N_{i=1}$ for each of the $N$ categories which are sampled from the base-classes dataset. 

\begin{figure*}[t]
\begin{center}
\includegraphics[scale=0.385]{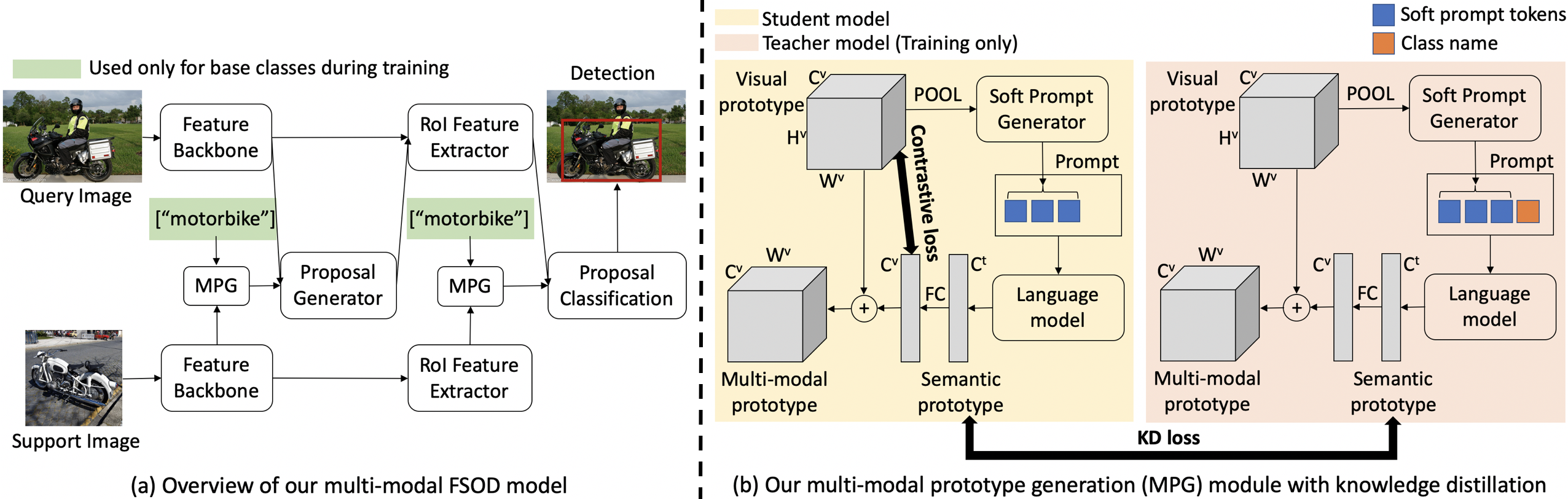}
\end{center}
\caption{(a) The overall architecture of our model for multi-modal FSOD. (b) The details of the multi-modal prototype generation module (MPG) with knowledge distillation, including the student model (left) and teacher model (right).}
\label{figure_6}
\end{figure*}

As shown in Fig.~\ref{figure_6}(a), we build our detection model using the siamese Faster R-CNN network, following~\cite{fan2020few,han2021meta}. It consists of two sequential stages: first proposal generation and then proposal classification:

\emph{Proposal Generation.} Given a query image and the support images of the category $i$, a shared feature backbone network (ResNet-101 till \texttt{res4} block~\cite{he2016deep}) is used to extract features of the query and support images as $f_p$ and $f_s$, respectively. Then, based on the support feature $f_s$, we generate the soft prompt tokens, which are then used to extract the class semantic prototype and multi-modal prototype $p_i$ with the proposed multi-modal prototype generation module (MPG). Class names are only used for base classes in MPG during training. Then, based on the proposal generation network in~\cite{fan2020few,han2021meta}, we generate class-specific proposals in the query image for the category $i$ using the multi-modal prototype $p_i$ and query feature $f_p$.

\emph{Proposal Classification.} Then, we use RoIAlign~\cite{he2017mask} and the \texttt{res5} block to extract the proposal features $f_p^{'}$ and the final support features $f_s^{'}$ respectively. Similarly, based on the support features $f_s^{'}$, we generate the multi-modal prototype $p_i^{'}$ using the proposed MPG module. Then, we use the pairwise matching network proposed in~\cite{fan2020few,han2021meta} to calculate the similarity between the proposal features and multi-modal prototype $p_i^{'}$ and also perform bbox regression to produce the final detection results.

\subsection{The Multi-modal Prototype Generation (MPG)}
\label{multimodal_prototype}

As in Fig.~\ref{figure_6}(b), we first extract few-shot visual prototypes $\{p_i^v\}^N_{i=1}$ and class semantic prototypes $\{\bar{p}_i\}^N_{i=1}$respectively, and then generate the multi-modal prototype $\{p_i\}^N_{i=1}$ by fusing the prototypes from the two modalities.

\textbf{Few-shot Visual Prototypes.} We calculate the average features of the $K$-shot support images as the visual prototype for each category, defined as,
\begin{equation}
p_i^v = \frac{1}{K}\textstyle{\sum}_{j=1}^{K}{F^v(I_i^j)}, \qquad p_i^v\in \mathbb{R}^{H^v*W^v*C^v}
\end{equation}
where $F^v$ is the visual feature extractor, and $\{I_i^j\}^K_{j=1}$ is the $K$-shot support images of the category $i$. $H^v$, $W^v$, and $C^v$ are the height, width, and channel numbers of the visual prototype $p_i^v$ respectively.

\textbf{Class Semantic Prototypes.} We use a pre-trained language model to extract class semantic prototypes. As shown in a recent work~\cite{radford2021learning}, designing proper prompt templates is crucial to fully exploit the pre-trained language model. However, previous works usually need heavy prompt engineering and ensembling~\cite{radford2021learning,li2022languagedriven,gu2022openvocabulary} which is sub-optimal, or needs additional few-shot fine-tuning to learn soft prompts~\cite{zhou2021learning}, which is prone to overfitting to the small training data. We argue that the few-shot support images include context information related to the category. Thus, \emph{we propose to learn a cross-modal soft prompt generator $G$ to generate the soft prompts, based on the few-shot visual examples}.

Different from previous prompt learning works~\cite{zhou2021learning,zhou2022conditional} which usually combine the soft prompt with different class names to obtain the final prompts. In the real world, it is usually hard to know the class names for rare classes. To address this problem, inspired by knowledge distillation~\cite{hinton2015distilling}, we propose to learn a student soft prompt generator without class names by transferring the knowledge from a teacher model with class names during the training over base classes. The teacher model is only used during training, and after training the student model is used for evaluation. 
\begin{align}
s_i^S &= G_S(\mathbf{POOL}(p_i^v)), \qquad s_i^S \in \mathbb{R}^{M*C^t} \\
\bar{p}_i^S &= F^t([s_i^S]), \qquad\qquad\quad\ \ \, \bar{p}_i^S \in \mathbb{R}^{C^t} \\
s_i^T &= G_T(\mathbf{POOL}(p_i^v)), \qquad s_i^T \in \mathbb{R}^{M*C^t} \\
\bar{p}_i^T &= F^t([s_i^T, E^t(t_i)]), \qquad \ \ \bar{p}_i^T \in \mathbb{R}^{C^t}
\end{align}
where $s_i^S$ and $s_i^T$ is the generated soft prompt for the category $i$ using the student model $G_S$ and teacher model $G_T$ respectively. $\bar{p}_i^S$ and $\bar{p}_i^T$ are the extracted class semantic prototypes using the pre-trained language model $F^t$. $\mathbf{POOL}$ is the spatial average pooling operation to convert $p_i^v$ into a vector with the dimension $C^v$. $M$ is the number of learnable prompt tokens, and $C^t$ is the dimension of token embedding, which is the same as the pre-trained token embedding $E^t$ in the language model $F^t$. We show in the experiment section, the results of an ablation study of using different numbers of learnable prompt tokens.
As for the soft prompt generator $G$, we use a simple fully-connected layer to convert the channel number of the input from $C^v$ to $M*C^t$. We empirically show that using this simple architecture leads to strong generalization ability for the few-shot novel classes during meta-testing, compared with using other complex networks, \eg, Transformer-based models~\cite{rao2021denseclip}.

\textbf{Multi-modal Prototypes.} we fuse the few-shot visual prototype $p_i^v$ and class semantic prototype $\bar{p}_i^S$/$\bar{p}_i^T$ for the multi-modal prototype $p_i^S$/$p_i^T$ using the feature fusion network $F$, which is defined as,
\begin{equation}
\begin{aligned}
& p_i^S = F(\bar{p}_i^S, p_i^v) = \mathbf{FC}(\bar{p}_i^S) + p_i^v, \quad p_i^S\in \mathbb{R}^{H^v*W^v*C^v} \\
& p_i^T = F(\bar{p}_i^T, p_i^v) = \mathbf{FC}(\bar{p}_i^T) + p_i^v, \quad p_i^T\in \mathbb{R}^{H^v*W^v*C^v}  \\
\end{aligned}
\end{equation}
where $\mathbf{FC}$ is a fully-connected layer to convert the channel number of the semantic prototype from $C^t$ to $C^v$, such that the converted semantic prototypes and the visual prototypes have the same channel number. We use the simple addition operation for multi-modal fusion.

The proposed MPG module is meta-learned over base classes, with the ability to prompt the pre-trained language model using few-shot visual examples, and dynamically fuse the visual and semantic prototypes. Our experiments confirm that the proposed module is stronger than other baselines without fine-tuning, \eg, manually designed prompt and shared soft prompt learned across classes.

\subsection{Training Objectives}
\label{loss}

\begin{figure}[t]
\begin{center}
\includegraphics[scale=0.36]{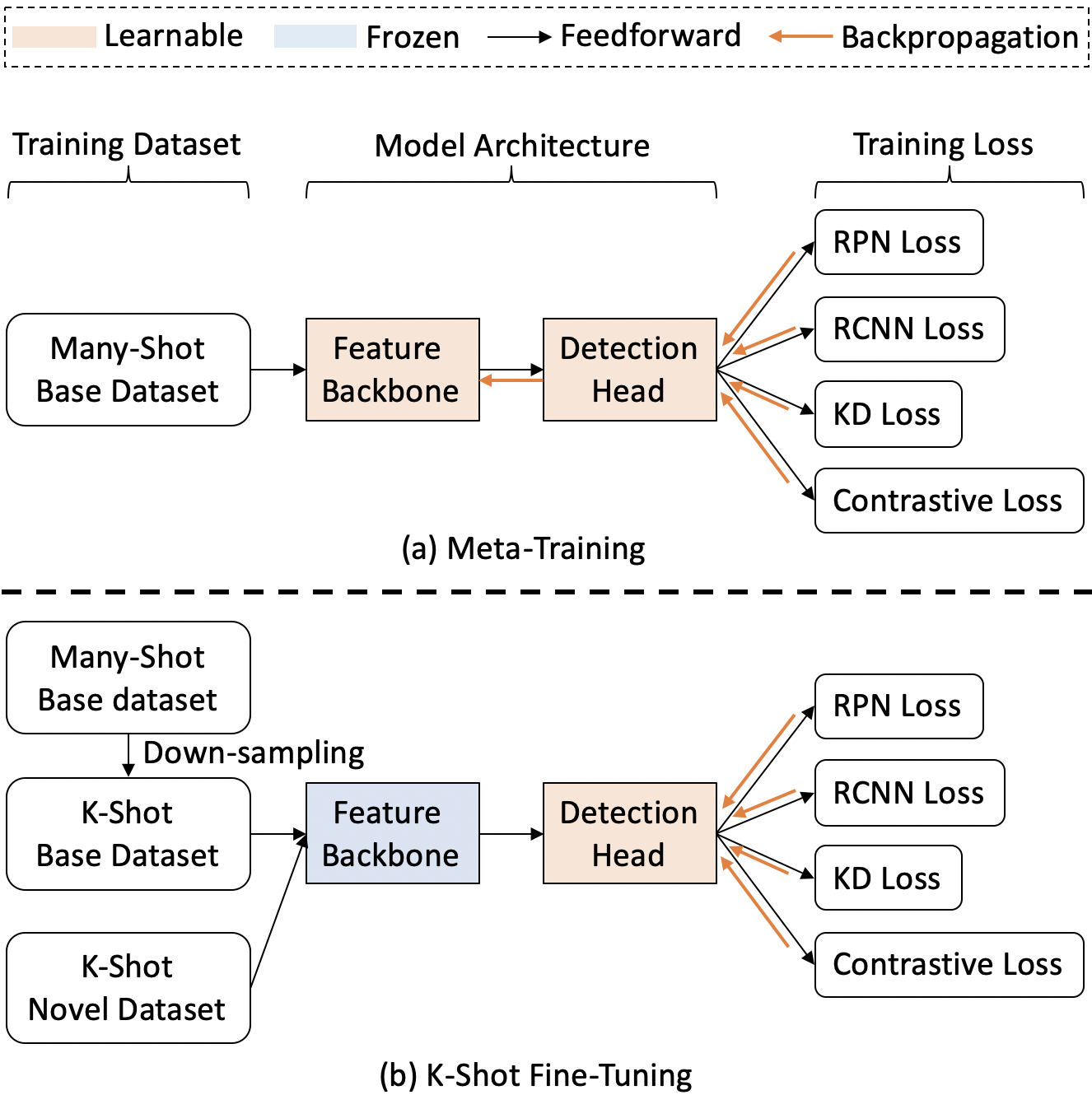} 
\end{center}
\caption{Meta-training vs. $K-$shot fine-tuning. We show only one branch of the model for simplicity.}
\label{figure_7}
\end{figure}

As shown in Fig.~\ref{figure_7}, we have two stages for model training, to fully exploit the data-abundant base dataset and the few-shot novel dataset for learning.

\subsubsection{Meta-training over Base Classes}

We sample multiple multi-modal FSOD learning tasks (a.k.a episodes) from the base-classes training data to simulate the few-shot learning scenarios of the novel classes. Each episode consists of query images with annotations, and few-shot support images together with the class names. The meta-learned model can be directly generalized to novel classes during meta-testing, without fine-tuning.

The training losses consist of the following parts: the binary classification loss and bbox regression loss in the proposal generation $L_{RPN}$ and proposal classification module $L_{RCNN}$ (following prior works~\cite{fan2020few,han2021meta}), the teacher-student knowledge distillation (KD) loss $L_{KD}$ and the visual-semantic contrastive loss $L_{C}$ in each of the two MPG modules. The training losses are defined as,
\begin{equation}
\label{loss_tot}
L_{TOT} = L_{RPN} + L_{RCNN} + L_{KD} + L_{C}
\end{equation}
where we follow the previous works~\cite{fan2020few,han2021meta} to implement the $L_{RPN}$ and $L_{RCNN}$ losses for the proposal generation and proposal classification modules respectively. $L_{KD}$ and $L_{C}$ are defined as follows.

\textbf{KD Loss.} To extract accurate semantic prototypes by the student model which does not class names, we enforce the class semantic prototypes extracted by the student and teacher model to be identical. We simply use the Euclidean distance to calculate the KD loss, which is defined as,
\begin{equation}
L_{KD} = \frac{1}{N}\sum_i || \bar{p}_i^S - \bar{p}_i^T ||_2.
\end{equation}

\textbf{Contrastive Loss.} We introduce a visual-semantic contrastive loss to encourage the alignment between vision and semantic feature space. During training, we use the student model to calculate the loss, which is to maximize the cosine similarity of corresponding visual and semantic prototypes, and minimize the cosine similarity of the incorrect pairs.
\begin{equation}
\begin{aligned}
L_{C} &= \frac{-1}{2N}\sum_i (\log \frac{\exp(p_i^v{\cdot}\mathbf{FC}(\bar{p}_i^T){/}\tau)}{\sum_j\exp(p_i^v{\cdot}\mathbf{FC}(\bar{p}_j^T){/}\tau)} \\
&+ \log \frac{\exp(\mathbf{FC}(\bar{p}_i^T){\cdot}p_i^v{/}\tau)}{\sum_j\exp(\mathbf{FC}(\bar{p}_i^T){\cdot}p_j^v{/}\tau)}),
\end{aligned}
\end{equation}
where $\tau$ is a temperature hyper-parameter.

\subsubsection{(Optional) Few-shot Fine-tuning over Novel Classes}

During meta-training, the model parameters are only learned using the base-classes dataset. We can further improve the model adaptation to novel classes after few-shot fine-tuning. To this end, for $k-$shot fine-tuning, we sample a small balanced dataset with both base classes and novel classes, where each class has exactly $k-$shot support images in the sampled dataset. Then we tune the meta-trained models using the sampled small dataset. To mitigate the potential overfitting issue during few-shot fine-tuning, we only tune the model parameters in the detection head and the deep feature backbone is frozen by default.

We use the same loss function as meta-training for fine-tuning, defined in Equation \ref{loss_tot}. The major difference between meta-training and few-shot fine-tuning is that we only use the base classes dataset during meta-training, while including the few-shot training data of novel classes during fine-tuning. Compared with the meta-trained models, few-shot fine-tuning can further improve model performance for novel classes after tuning the model parameters with a few training examples from novel classes.

\section{Experimental Results}

\subsection{Datasets}

We evaluated our model on two widely used FSOD benchmarks, the MSCOCO~\cite{lin2014microsoft} and PASCAL VOC dataset~\cite{everingham2010pascal} following the evaluation protocol defined in ~\cite{wang2020few}. 

\textbf{PASCAL VOC.} Following previous works in~\cite{kang2019few,wang2020few}, we have three random partitions of base and novel categories. In each partition, the twenty PASCAL VOC categories are split into fifteen base classes and five novel classes. 
We have the exact same few-shot images for model training/testing as~\cite{wang2020few,Sun_2021_CVPR}, and report AP50 results under shots 1, 2, 3, 5, and 10. We report both meta-testing results and few-shot fine-tuning results following~\cite{han2021meta,Han_2021_ICCV}.

\begin{table*}[t]
    \centering
    \caption{Ablation study on the major model designs.}
    \addtolength{\tabcolsep}{-0.1cm}
    \adjustbox{width=\linewidth}{
    \begin{tabular}{c|l|c|c|ccc|ccc|ccc}
    \toprule
    & \multirow{2}{*}{Method} & \multirow{2}{*}{Language} & \multirow{2}{*}{Vision} & \multicolumn{3}{c|}{1-shot} & \multicolumn{3}{c|}{2-shot} & \multicolumn{3}{c}{10-shot}\\
    & & & & AP & AP50 & AP75 & AP & AP50 & AP75 & AP & AP50 & AP75 \\ \midrule
    \multicolumn{13}{c}{\textbf{Meta-training the model on base classes, and meta-testing on novel classes}} \\ \midrule
    (a) & Using hard prompt of ``[class\_name]'' & \checkmark & & 2.2 & 4.1 & 2.1  & 2.2 & 4.1 & 2.1  & 2.2 & 4.1 & 2.1 \\
    (b) & Using hard prompt of ``a photo of [class\_name]'' & \checkmark &  & 3.0 & 5.4 & 3.0  & 3.0 & 5.4 & 3.0  & 3.0 & 5.4 & 3.0 \\
    (c) & Using hard prompts ensemble & \checkmark & & 3.4 & 6.5 & 3.3   & 3.4 & 6.5 & 3.3   & 3.4 & 6.5 & 3.3 \\
    (d) & Using shared soft prompt across classes \cite{zhou2021learning} & \checkmark & & 1.6 & 3.0 & 1.5  & 1.6 & 3.0 & 1.5  & 1.6 & 3.1 & 1.6 \\ 
    (e) & Our teacher model w/ class name & \checkmark & & 3.7 & 6.6 & 3.8  & 4.1 & 7.4 & 4.1  & 4.7 & 8.3 & 4.8 \\
    (f) & Our student model w/o class name & \checkmark & & 3.7 & 6.6 & 3.7  & 3.9 & 7.2 & 3.8  & 4.8 & 8.6 & 4.9 \\ \midrule
    (g) & Using few-shot model only & & \checkmark & {5.0} & {10.2} & {4.6}    & {7.0} & {13.5} & {6.4}    & {9.7} & {18.5} & {9.0} \\ \midrule
    (h) & Our MM-FSOD & \checkmark & \checkmark &  \textbf{5.6} & {11.0} & \textbf{5.2}   & {7.9} & {15.3} & \textbf{7.4} & {10.8} & {20.5} & {10.2} \\ \midrule
    \multicolumn{13}{c}{\textbf{Fine-tuning the model on novel classes, and testing on novel classes}} \\ \midrule
    (i) & Our MM-FSOD & \checkmark & \checkmark & {5.4} & \textbf{11.3} & {4.8}    & \textbf{8.4} & \textbf{17.2} & {7.0}   & \textbf{13.3} & \textbf{27.5} & \textbf{11.2} \\
    \bottomrule
    \end{tabular}}
\label{tab:main_ablation}
\end{table*}

\textbf{MSCOCO.} We use the twenty PASCAL VOC categories as novel classes and the remaining sixty categories are base classes. We have the exact same few-shot images for model training/testing as~\cite{wang2020few,Sun_2021_CVPR}, and report the detection accuracy AP/AP50/AP75 under shots 1, 2, 3, 5, 10 and 30 following~\cite{Qiao_2021_ICCV,Han_2021_ICCV,wang2020few}. We report both meta-testing results and few-shot fine-tuning results following~\cite{han2021meta,Han_2021_ICCV}.

We use the MSCOCO dataset under 1/2/10-shots for the ablation study in Section \ref{ablation}, and report the full results on the two FSOD benchmarks in Section \ref{sotas}.

\subsection{Implementation Details}
\label{imple_detail}

We implemented our model based on the previous meta-learning-based FSOD works~\cite{fan2020few,han2021meta}, and followed most of the model designs and hyperparameters in their works. The hyperparameter temperature $\tau=0.01$. We would like to emphasize that we use ResNet-101 as the image feature extractor, which is the exact same as most of the previous FSOD methods. 
For the text encoder, we use the text-only pre-trained language model BERT by default, and use other pre-trained language models achieve similar performance (\eg, the CLIP-ResNet101 text encoder). The parameters of the text encoder are fixed during training. 
In this way, the only difference between our multi-modal FSOD models and the traditional vision-only FSOD models is that we use the additional class semantic information extracted from a strong pre-trained language model to develop our models. Thus, the performance gain only comes from the newly-introduced class semantic information.

For each episode during meta-training, we sample a 2-way 30-shot support set for each query image. Specifically, a positive and a negative support class are randomly selected for each query image. The positive class indicates that it appears in the query image, while the negative class does not. After meta-training, our model is tested over unseen novel classes during meta-testing. After meta-learning, we fine-tune the model over novel classes. During fine-tuning, the feature backbone is fixed, and we only tune the detection head using few-shot training data.

Specifically, we start with the pre-trained models in \cite{fan2020few} with the improved feature fusion networks proposed in \cite{han2021meta}. For meta-training on the MSCOCO dataset, we use the SGD optimizer with an initial learning rate of 0.001, a momentum of 0.9, a weight decay of 0.0001, and a batch size of 8. The learning rate is divided by 10 after 15,000 iterations. The total number of training iterations is 20,000. Similarly, for meta-training on the VOC dataset, we use the same hyper-parameters as on the MSCOCO dataset except using only half of the training iterations.

For few-shot fine-tuning, we use the SGD optimizer with an initial learning rate of 0.002, a momentum of 0.9, a weight decay of 0.0001, and a batch size of 8. The difference between meta-training is that we use much smaller training iterations for fine-tuning, and the feature backbone is frozen. The learning rate is divided by 10 after 2,000 iterations, and the total number of training iterations is 3,000. 

\subsection{Ablation Study}
\label{ablation}

We conducted comprehensive ablation studies on the MSCOCO dataset to verify the effectiveness of the model designs and hyperparameter selections as follows.

\textbf{Effectiveness of our meta-learning-based cross-modal prompting.} We compare different prompting methods in Table~\ref{tab:main_ablation} (a-f), including hard prompt engineering and ensembling, learnable soft prompt shared across classes, and our proposed method. We have the following three findings: \textbf{(1)}. Directly using class names as the prompt is usually sub-optimal. This is because, during CLIP pre-training, image-caption pairs are used for modal training. Therefore, prompt design with proper context is important for downstream tasks. Using the prompts in Table~\ref{tab:main_ablation} (b) and (c) as suggested by the original CLIP paper~\cite{radford2021learning}, we can observe a huge relative improvement, compared with Table~\ref{tab:main_ablation} (a). \textbf{(2)}. Following~\cite{zhou2021learning}, we attempt to learn shared soft prompts across base classes during meta-training as meta-knowledge, which can be generalized to novel classes during meta-testing. However, as shown in Table~\ref{tab:main_ablation} (d), the results are even worse than Table~\ref{tab:main_ablation} (a). The reason is that, in~\cite{zhou2021learning}, there are no unseen classes during testing and all classes are used to train the shared soft prompt. Moreover, the learned prompt is fixed during testing and thus may not be suitable for unseen classes. \textbf{(3)}. Our proposed meta-learning-based cross-modal prompting method does not learn the soft prompt as meta-knowledge, and instead learns the soft prompt prediction module as meta-knowledge, conditioned on the few-shot visual examples. Therefore, our method, shown in Table~\ref{tab:main_ablation} (e) and (f), can dynamically generate proper soft prompts for novel classes during meta-testing, and shows much improved performance compared with the method in Table~\ref{tab:main_ablation} (d). Using knowledge distillation can learn competitive student models without class names compared with the teacher model. In addition, our method has a similar performance as prompt ensembling under 1-shot, and our performance steadily improves with more few-shot visual examples. All these results validate the effectiveness of our meta-learning-based cross-modal prompting method.

\textbf{Effectiveness of our multi-modal prototype fusion.} The language-only and vision-only results are shown in Table~\ref{tab:main_ablation} (f) and Table~\ref{tab:main_ablation} (g) respectively. We see that our language-only model performs slightly lower compared with the 1-shot vision-only model. Using more shots, the vision-only model can be boosted largely. This shows that directly applying the language model to the MSCOCO dataset is very challenging because of the small number of classes~\cite{gu2022openvocabulary}. Considering the complementarity of visual and semantic information, our final model MM-FSOD, in Table~\ref{tab:main_ablation} (h), achieves consistent improvement across all shots, compared with any of the single-modality models. 

Furthermore, our model can be improved after few-shot fine-tuning, especially with large shots, \eg, 10-shot. The meta-learning-only method has better results under low shots, \eg, 1-shot, similar to~\cite{han2021meta,Han_2021_ICCV}.

\begin{table}[t]
    \centering
    \caption{Ablation study on the number of prompt tokens for both of the teacher and student models.} 
    \addtolength{\tabcolsep}{-0.05cm}
    \adjustbox{width=0.9\linewidth}{
    \begin{tabular}{l|ccc|ccc}
    \toprule
    \multirow{2}{*}{\#tokens} & \multicolumn{3}{c|}{2-shot} & \multicolumn{3}{c}{10-shot}\\
    & AP & AP50 & AP75 & AP & AP50 & AP75 \\ \midrule
    2 & 3.4 & 6.3 & 3.2   & 4.0 & 7.5 & 3.9 \\
    4 & 3.8 & 7.0 & 3.6   & 4.5 & 8.0 & 4.6 \\
    8 & \textbf{3.9} & \textbf{7.2} & \textbf{3.8}   & \textbf{4.8} & \textbf{8.6} & \textbf{4.9} \\
    16 & 3.8 & 7.1 & 3.7  & 4.6 & 8.3 & 4.6 \\
    \bottomrule
    \end{tabular}}
\label{tab:number_of_token}
\end{table}

\begin{table}[t]
    \centering
    \caption{Ablation study on the position of the soft prompt in the teacher model. x: prompt token.}
    \addtolength{\tabcolsep}{-0.08cm}
    \adjustbox{width=\linewidth}{
    \begin{tabular}{l|ccc|ccc}
    \toprule
    \multirow{2}{*}{Prompt Position} & \multicolumn{3}{c|}{2-shot} & \multicolumn{3}{c}{10-shot}\\
    & AP & AP50 & AP75 & AP & AP50 & AP75 \\ \midrule
    $\mathrm{xx}\left[\mathrm{CLASS\_NAME}\right]$ & \textbf{3.9} & \textbf{7.2} & \textbf{3.8}   & \textbf{4.8} & \textbf{8.6} & \textbf{4.9} \\
    $\left[\mathrm{CLASS\_NAME}\right]\mathrm{xx}$ & 3.7 & 7.0 & 3.6           & 4.6 & 8.3 & 4.7 \\
    $\mathrm{x}\left[\mathrm{CLASS\_NAME}\right]\mathrm{x}$ & 3.6 & 6.8 & 3.6  & 4.5 & 8.4 & 4.6 \\
    \bottomrule
    \end{tabular}
    }
\label{tab:position}
\end{table}

\textbf{The ablation study on the number of learnable soft prompt tokens and the position.}
We show in Table~\ref{tab:number_of_token}, that the performance improves when increasing the number of learnable tokens from two to eight. However, the performance is saturated and the improvements diminish if further increasing the context length. Therefore, we use eight soft prompt tokens for both of the teacher and student models by default. Besides, as shown in Table~\ref{tab:position}, we empirically find that putting the prompt tokens before the class name token in the teacher model, has slightly better results. These empirical findings generalize well to the VOC dataset. 

\begin{table}[t]
    \centering
    \caption{Ablation study on the soft prompt generator $G$.}
    \addtolength{\tabcolsep}{-0.05cm}
    \adjustbox{width=\linewidth}{
    \begin{tabular}{l|ccc|ccc}
    \toprule
    \multirow{2}{*}{Generator G} & \multicolumn{3}{c|}{2-shot} & \multicolumn{3}{c}{10-shot}\\
    & AP & AP50 & AP75 & AP & AP50 & AP75 \\ \midrule
    One MLP layer & \textbf{3.9} & \textbf{7.2} & \textbf{3.8}   & \textbf{4.8} & \textbf{8.6} & \textbf{4.9} \\
    Two MLP layers & 3.6 & 6.8 & 3.6    & 4.4 & 8.1 & 4.6 \\
    Pre-Transformer & 3.3 & 6.0 & 3.4   & 4.0 & 6.9 & 4.4 \\
    Post-Transformer & 1.9 & 4.0 & 1.9  & 2.6 & 4.9 & 2.8 \\
    \bottomrule
    \end{tabular}}
\label{tab:generator}
\end{table}

\begin{table*}[t]
\centering
\footnotesize
\caption{{Few-shot object detection performance (AP50) on the PASCAL VOC dataset, with both meta-testing and fine-tuning results.}}
\addtolength{\tabcolsep}{-0.1cm}
\adjustbox{width=\linewidth}{
\begin{tabular}{l|c|ccccc|ccccc|ccccc}
\toprule
\multirow{2}{*}{Method} & \multirow{2}{*}{Venue} & \multicolumn{5}{c|}{Novel Set 1} & \multicolumn{5}{c|}{Novel Set 2} & \multicolumn{5}{c}{Novel Set 3} \\ 
&  &1     & 2     & 3    & 5    & 10   & 1     & 2     & 3    & 5    & 10   & 1     & 2     & 3    & 5    & 10   \\ \midrule
\multicolumn{17}{c}{\textbf{Meta-training the model on base classes, and meta-testing on novel classes}} \\ \midrule
Fan et al.~\cite{fan2020few} & CVPR 2020 & 32.4 & 22.1 & 23.1 & 31.7 & 35.7    & 14.8 & 18.1 & 24.4 & 18.6 & 19.5 &    25.8 & 20.9 & 23.9 & 27.8 & 29.0 \\
QA-FewDet~\cite{Han_2021_ICCV} & ICCV 2021 & {41.0} & {33.2} & {35.3} & {47.5} & {52.0}   & {23.5} & {29.4} & {37.9} & {35.9} & {37.1}   & {33.2} & {29.4} & {37.6} & {39.8} & {41.5} \\
Meta Faster R-CNN~\cite{han2021meta} & AAAI 2022 & {40.2} & {30.5} & {33.3} & {42.3} & {46.9}   & {26.8} & {32.0} & {39.0} & \textbf{37.7} & {37.4}   & {34.0} & {32.5} & {34.4} & \textbf{42.7} & {44.3} \\ 
MM-FSOD (Ours) & This work & \textbf{42.5} & \textbf{41.2} & \textbf{41.6} & \textbf{48.0} & \textbf{53.4}  & \textbf{30.5} & \textbf{34.0} & \textbf{39.3} & 36.8 & \textbf{37.6} & \textbf{39.9} & \textbf{37.0} & \textbf{38.2} & {42.5} & \textbf{45.6} \\ \midrule
\multicolumn{17}{c}{\textbf{Fine-tuning the model on novel classes, and testing on novel classes}} \\ \midrule
FSRW~\cite{kang2019few}  & ICCV 2019 & 14.8  & 15.5  & 26.7 & 33.9 & 47.2 & 15.7  & 15.3  & 22.7 & 30.1 & 40.5 & 21.3  & 25.6  & 28.4 & 42.8 & 45.9 \\ 
MetaDet~\cite{wang2019meta} & ICCV 2019 & 18.9 & 20.6 & 30.2 & 36.8 & 49.6 & 21.8 & 23.1 & 27.8 & 31.7 & 43.0 & 20.6 & 23.9 & 29.4 & 43.9 & 44.1 \\ 
Meta R-CNN~\cite{yan2019meta} & ICCV 2019 & 19.9 & 25.5 & 35.0 & 45.7 & 51.5 & 10.4 & 19.4 & 29.6 & 34.8 & 45.4 & 14.3 & 18.2 & 27.5 & 41.2 & 48.1 \\ 
TFA w/ fc~\cite{wang2020few} & ICML 2020 & {36.8} & {29.1} & {43.6} & {55.7} & {57.0} & {18.2} & {29.0} & {33.4} & {35.5} & {39.0} & {27.7} & {33.6} & {42.5} & {48.7} & {50.2}\\
TFA w/ cos~\cite{wang2020few} & ICML 2020 & 39.8 & 36.1 & 44.7 & 55.7 & 56.0 & 23.5 & 26.9 & 34.1 & 35.1 & 39.1 & 30.8 & 34.8 & 42.8 & 49.5 & 49.8 \\ 
Xiao et al.~\cite{xiao2020few} & ECCV 2020 & 24.2 & 35.3 &  42.2 &  49.1 &  57.4 & 21.6 & 24.6 &  31.9 &  37.0 &  45.7 & 21.2 &  30.0 &  37.2 &  43.8 &  49.6 \\
MPSR~\cite{wu2020multi} & ECCV 2020 & 41.7 & 42.5 & 51.4 & 55.2 & 61.8 & 24.4 & 29.3 & 39.2 & 39.9 & 47.8 & 35.6 & 41.8 & 42.3 & 48.0 & 49.7 \\ 
Fan et al.~\cite{fan2020few} & CVPR 2020 & 37.8 & 43.6 & 51.6 & 56.5 & 58.6    & 22.5 & 30.6 & 40.7 & 43.1 & 47.6    & 31.0 & 37.9 & 43.7 & 51.3 & 49.8\\ 
SRR-FSD~\cite{Zhu_2021_CVPR} & CVPR 2021 & \textbf{47.8} & 50.5 & 51.3 & 55.2 & 56.8    & \textbf{32.5} & 35.3 & 39.1 & 40.8 & 43.8    & 40.1 & 41.5 & 44.3 & 46.9 & 46.4 \\
TFA + Halluc~\cite{Zhang_2021_CVPR} & CVPR 2021 & 45.1 & 44.0 & 44.7 & 55.0 & 55.9   & 23.2 & 27.5 & 35.1 & 34.9 & 39.0   & 30.5 & 35.1 & 41.4 & 49.0 & 49.3 \\
CoRPNs + Halluc~\cite{Zhang_2021_CVPR} & CVPR 2021 & 47.0 & 44.9 & 46.5 & 54.7 & 54.7   & 26.3 & 31.8 & 37.4 & 37.4 & 41.2   & 40.4 & 42.1 & 43.3 & 51.4 & 49.6 \\
FSCE~\cite{Sun_2021_CVPR} & CVPR 2021 & 44.2 & 43.8 & 51.4 & 61.9 & 63.4    & 27.3 & 29.5 & 43.5 & 44.2 & 50.2    & 37.2 & 41.9 & 47.5 & 54.6 & 58.5 \\
FSOD$^{up}$~\cite{Wu_2021_ICCV} & ICCV 2021 & 43.8 & 47.8 & 50.3 & 55.4 & 61.7   & 31.2 & 30.5 & 41.2 & 42.2 & 48.3   & 35.5 & 39.7 & 43.9 & 50.6 & 53.5 \\
QA-FewDet~\cite{Han_2021_ICCV} & ICCV 2021 & {42.4} & {51.9} & {55.7} & {62.6} & {63.4} & {25.9} & \textbf{37.8} & {46.6} & {48.9} & {51.1} & 35.2 & {42.9} & {47.8} & {54.8} & {53.5} \\
Meta Faster R-CNN~\cite{han2021meta} & AAAI 2022 & 43.0 & {54.5} & {60.6} & \textbf{66.1} & {65.4}   & 27.7 & {35.5} & {46.1} & {47.8} & \textbf{51.4}   & {40.6} & {46.4} & {53.4} & \textbf{59.9} & {58.6}\\
MM-FSOD (Ours) & This work & 46.8 & \textbf{55.2} & \textbf{61.3} & 65.8 & \textbf{66.0}    & 31.2 & 37.3 & \textbf{46.9} & \textbf{49.2} & 51.2    & \textbf{41.3} & \textbf{47.2}  & \textbf{53.8} & 59.6 & \textbf{59.3} \\ 
\bottomrule
\end{tabular}}
\label{tab:main_voc}
\end{table*}

\begin{table}[t]
    \centering
    \caption{Ablation study on the multi-modal feature fusion operation in the MPG module.}
    \addtolength{\tabcolsep}{-0.05cm}
    \adjustbox{width=\linewidth}{
    \begin{tabular}{l|ccc|ccc}
    \toprule
    \multirow{2}{*}{Fusion} & \multicolumn{3}{c|}{2-shot} & \multicolumn{3}{c}{10-shot}\\
    & AP & AP50 & AP75 & AP & AP50 & AP75 \\ \midrule
    Addition & \textbf{7.9} & \textbf{15.3} & \textbf{7.4} & \textbf{10.8} & \textbf{20.5} & \textbf{10.2} \\
    Multiplication & 7.2 & 13.8 & 6.6  & 9.4 & 18.2 & 8.6 \\
    Concatenation & 6.3 & 12.7 & 5.8  & 9.0 & 17.3 & 8.2 \\ 
    \bottomrule
    \end{tabular}}
\label{tab:Fusion}
\end{table}

\begin{table}[t]
    \centering
    \caption{Ablation study on applying our MPG to the detection model. Only the vision model is used if not marked.}
    \addtolength{\tabcolsep}{-0.05cm}
    \adjustbox{width=\linewidth}{
    \begin{tabular}{c|c|ccc|ccc}
    \toprule
    \multirow{2}{*}{RPN} & \multirow{2}{*}{RCNN} & \multicolumn{3}{c|}{2-shot} & \multicolumn{3}{c}{10-shot}\\
     &  & AP & AP50 & AP75 & AP & AP50 & AP75 \\ \midrule
    & & {7.0} & {13.5} & {6.4}    & {9.7} & {18.5} & {9.0} \\
    & \checkmark & {7.8} & {15.1} & {7.2}    & {10.5} & {20.1} & {9.8} \\
    \checkmark & \checkmark & \textbf{7.9} & \textbf{15.3} & \textbf{7.4} & \textbf{10.8} & \textbf{20.5} & \textbf{10.2} \\ 
    \bottomrule
    \end{tabular}}
\label{tab:RPN_RCNN}
\end{table}

\begin{table*}[t]
\centering
\footnotesize
\caption{Few-shot object detection performance on the MSCOCO dataset, with both meta-testing and fine-tuning results.}
\addtolength{\tabcolsep}{-0.15cm}
\adjustbox{width=\linewidth}{
\begin{tabular}{l|ccc|ccc|ccc|ccc|ccc|ccc}
\toprule
&\multicolumn{3}{c|}{1-shot} & \multicolumn{3}{c|}{2-shot} & \multicolumn{3}{c|}{3-shot} &\multicolumn{3}{c|}{5-shot} & \multicolumn{3}{c|}{10-shot} & \multicolumn{3}{c}{30-shot} \\
Method & AP & AP50 & AP75 & AP & AP50 & AP75 & AP & AP50 & AP75 & AP & AP50 & AP75 & AP & AP50 & AP75 & AP & AP50 & AP75 \\ \midrule
\multicolumn{19}{c}{\textbf{Meta-training the model on base classes, and meta-testing on novel classes}} \\ \midrule
Fan et al.~\cite{fan2020few} & 4.0 & 8.5 & 3.5  & 5.4 & 11.6 & 4.6    & 5.9 & 12.5 & 5.0   & 6.9 & 14.3 & 6.0    & 7.6 & 15.4 & 6.8  & 8.9 & 17.8 & 8.0 \\
QA-FewDet~\cite{Han_2021_ICCV} &  {5.1} & {10.5} & {4.5}    & {7.8} & \textbf{16.4} & {6.6}    & {8.6} & {17.7} & {7.5}    & {9.5} & {19.3} & {8.5}  & {10.2} & {20.4} & {9.0}   & {11.5} & \textbf{23.4} & {10.3} \\ 
Meta Faster R-CNN~\cite{han2021meta} & {5.0} & {10.2} & {4.6}    & {7.0} & {13.5} & {6.4}    & {8.4} & {16.5} & {7.4}    & {9.3} & {18.1} & {8.3}   & {9.7} & {18.5} & {9.0} & {11.3} & {21.2} & {10.6} \\
MM-FSOD (Ours) & \textbf{5.6} & \textbf{11.0} & \textbf{5.2}   & \textbf{7.9} & 15.3 & \textbf{7.4}  & \textbf{9.4} & \textbf{18.3} & \textbf{8.9}   & \textbf{10.5} & \textbf{19.7} & \textbf{9.5}  & \textbf{10.8} & \textbf{20.5} & \textbf{10.2}   & \textbf{12.3} & 22.8 & \textbf{11.8} \\ 
\midrule
\multicolumn{18}{c}{\textbf{Fine-tuning the model on novel classes, and testing on novel classes}} \\ \midrule
FSRW\small{~\cite{kang2019few}}  & {--} & {--} & {--} & {--} & {--} & {--} & {--} & {--} & {--} &\;{--} & {--} & {--} & 5.6 & 12.3 & 4.6 & 9.1 & 19.0 & 7.6 \\ 
MetaDet\small{~\cite{wang2019meta}} & {--} & {--} & {--} & {--} & {--} & {--} & {--} & {--} & {--} &\;{--} & {--} & {--} & 7.1 & 14.6 & 6.1 & 11.3 & 21.7 & 8.1 \\
Meta R-CNN~\cite{yan2019meta} & {--} & {--} & {--} & {--} & {--} & {--} & {--} & {--} & {--} &\;{--} & {--} & {--} & {8.7} & 19.1 & {6.6} & {12.4} & 25.3 & {10.8} \\
TFA w/ fc~\cite{wang2020few} & 2.9 & 5.7 & 2.8   & 4.3 & 8.5 & 4.1    & 6.7 & 12.6 & 6.6   & 8.4 & 16.0 & 8.4    & 10.0 & 19.2 & 9.2    & 13.4 & 24.7 & 13.2 \\
TFA w/ cos~\cite{wang2020few} & 3.4 & 5.8 & 3.8   & 4.6 & 8.3 & 4.8    & 6.6 & 12.1 & 6.5   & 8.3 & 15.3 & 8.0     & 10.0 & 19.1 & 9.3    & 13.7& 24.9 & 13.4 \\
Xiao et al.~\cite{xiao2020few} & 3.2 & 8.9 & 1.4  & 4.9 & 13.3 & 2.3  & 6.7 & {18.6} & 2.9  & 8.1 & 20.1 & 4.4 & 10.7 & {25.6} & 6.5 & 15.9 & 31.7 & 15.1 \\
MPSR~\cite{wu2020multi}    & {2.3} & {4.1} & {2.3}    & {3.5} & {6.3} & {3.4}    & {5.2} & {9.5} & {5.1}   &{6.7} & {12.6} & {6.4}    & {9.8} & {17.9} & {9.7}   & {14.1} & {25.4} & {14.2} \\
Fan et al.~\cite{fan2020few} & 4.2 & 9.1 & 3.0   & 5.6 & 14.0 & 3.9   & 6.6 & 15.9 & 4.9   & 8.0 & 18.5 & 6.3   & 9.6 & 20.7 & 7.7    & 13.5 & 28.5 & 11.7 \\
SRR-FSD~\cite{Zhu_2021_CVPR} & {--} & {--} & {--} & {--} & {--} & {--} & {--} & {--} & {--} &\;{--} & {--} & {--} & 11.3 & 23.0 & 9.8  & 14.7 & 29.2 & 13.5\\
TFA + Halluc~\cite{Zhang_2021_CVPR} & 4.4 & 7.5 & \textbf{4.9}    & 5.6 & 9.9 & 5.9   & 7.2 & 13.3 & 7.4 & {--} & {--} & {--} & {--} & {--} & {--} & {--} & {--} & {--} \\
CoRPNs + Halluc~\cite{Zhang_2021_CVPR} & 3.8 & 6.5 & 4.3    & 5.0 & 9.0 & 5.2   & 6.9 & 12.6 & 7.0 & {--} & {--} & {--} & {--} & {--} & {--} & {--} & {--} & {--} \\
FSCE~\cite{Sun_2021_CVPR} & {--} & {--} & {--} & {--} & {--} & {--} & {--} & {--} & {--} &\;{--} & {--} & {--} & 11.9 & - & 10.5   & 16.4 & - & \textbf{16.2} \\
FSOD$^{up}$~\cite{Wu_2021_ICCV} & {--} & {--} & {--} & {--} & {--} & {--} & {--} & {--} & {--} &\;{--} & {--} & {--} & 11.0 & - & 10.7 & 15.6 & - & 15.7 \\
QA-FewDet~\cite{Han_2021_ICCV} & {4.9} & {10.3} & {4.4}  & {7.6} & {16.1} & {6.2}    & {8.4} & 18.0 & {7.3}   & {9.7} & {20.3} & {8.6}   & {11.6} & 23.9 & {9.8} & {16.5} & {31.9} & {15.5} \\
Meta Faster R-CNN~\cite{han2021meta} & {5.1} & {10.7} & {4.3}  & {7.6} & {16.3} & {6.2}    & {9.8} & {20.2} & {8.2}   & {10.8} & {22.1} & {9.2}   & {12.7} & {25.7} & {10.8} & {16.6} & {31.8} & {15.8} \\
MM-FSOD (Ours) & \textbf{5.4} & \textbf{11.3} & 4.8    & \textbf{8.4} & \textbf{17.2} & \textbf{7.0}   & \textbf{10.5} & \textbf{21.4} & \textbf{8.6}   & \textbf{11.4} & \textbf{23.8} & \textbf{9.3}    & \textbf{13.3} & \textbf{27.5} & \textbf{11.2}   & \textbf{17.2} & \textbf{33.3} & {16.0} \\ 
\bottomrule
\end{tabular}}
\label{tab:main_coco}
\end{table*}

\textbf{The comparison of different soft prompt generation networks.}
We compare different model architectures for our soft prompt generation module in Table~\ref{tab:generator}, including multiple MLP layers, and the transformer-based network in~\cite{rao2021denseclip}. Using the simplest one-layer MLP network has the best results. When more MLP layers are used, the performance decreases due to the potential overfitting issue with more parameters. In~\cite{rao2021denseclip}, two vision-to-language prompting strategies are proposed. In pre-model prompting, a transformer decoder with learnable queries is used to extract visual context, which is also used as the soft prompt. In post-model prompting, a shared soft prompt is first learned across classes, similar to the method in Table~\ref{tab:main_ablation} (d), and then a transformer decoder is used to refine the text features with the visual clues. However, in~\cite{rao2021denseclip}, all classes are used to train the transformer networks and there are no unseen classes during testing. Moreover, the learnable queries in the pre-model prompting and the shared soft prompt learned in the post-model prompting are fixed during meta-testing, which may not be suitable for unseen classes. In contrast, our proposed meta-learning-based cross-modal prompting method can generate dynamic soft prompts for novel classes during meta-testing, and is much simpler compared with pre-model prompting. This explains the lower meta-testing results using~\cite{rao2021denseclip}, compared with our method.

\textbf{The comparison of different multi-modal fusion operations.} We compare different multi-modal fusion operations in Table~\ref{tab:Fusion}, including addition, multiplication and concatenation. The addition achieves the best performance. This is because the addition operation works as a residual connection and can largely preserve the advantages from the two modalities, while multiplication can hardly achieve this. Although the concatenation can also preserve the knowledge from the two modalities, it yields the worst results. This is because it needs an additional MLP layer to decrease the number of the concatenated channels to the same as the query features. This avoids the principle in siamese networks that each branch should go through the exact same number of learnable layers, such that the final features of the two branches are in the same feature space, especially for the visual features. Therefore, we use the addition for multi-modal fusion in our model.

\textbf{Effectiveness of applying our MPG module to the detection model.} 
We show in Table~\ref{tab:RPN_RCNN} the results of applying our MPG module to the proposal generation and classification module. Applying our MPG module to the proposal classification module brings the most gain, because the extracted multi-modal prototype is directly used to produce the final detection. Using our MPG module for proposal generation can slightly improve the quality of the generated proposals. Thus, we have two MPG modules in our model.

\begin{figure*}[t]
\begin{center}
\includegraphics[scale=1.22]{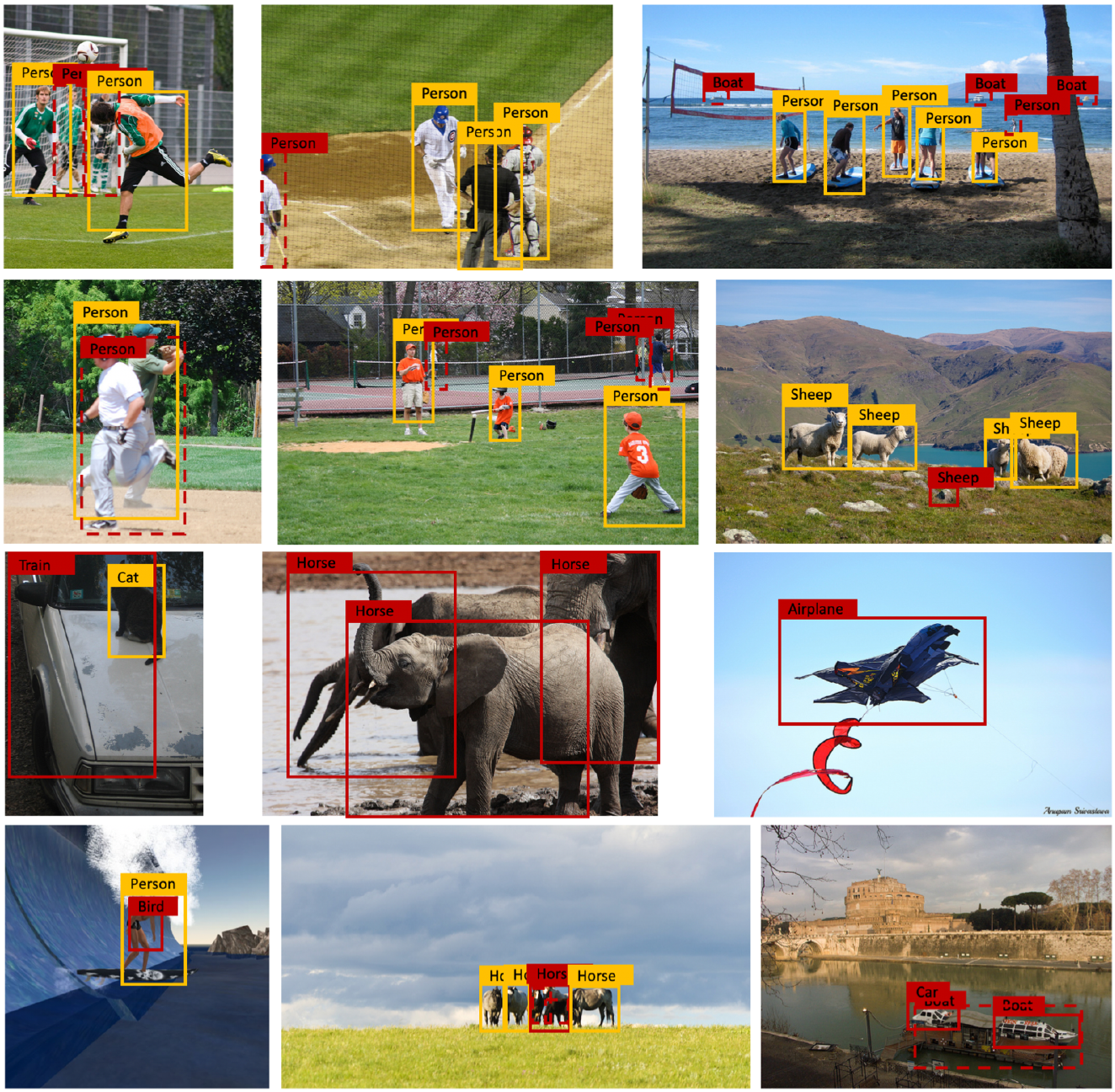} 
\end{center}
\caption{Visualization of detection results and the failure cases. We use the 30-shot fine-tuning model trained on the MSCOCO dataset, and the test images are from the MSCOCO validation dataset. Our model can achieve fairly good detection results on the challenging MSCOCO dataset. Typical failure cases include misclassification between confusing categories, missing objects especially the small objects, and etc. Future work can improve small object detection and the ability of few-shot classification. Yellow boxes indicate true positive detections. Red solid boxes indicate false positives (usually mis-classified detections). Red dashed boxes indicate false negatives (usually missing boxes).}
\label{figure_8}
\end{figure*}

\subsection{Comparison with the State-of-the-arts (SOTAs)}
\label{sotas}
We show in Tables~\ref{tab:main_voc} and \ref{tab:main_coco} the comparison of our proposed MM-FSOD with the other methods using both meta-learning-only and after fine-tuning, on PASCAL VOC and MSCOCO FSOD benchmarks respectively. 

First, only few methods~\cite{han2021meta,Han_2021_ICCV} report the meta-testing results. 
We argue that meta-testing is an important indicator to measure the generalization ability of the models with unseen classes. 
Another benefit of meta-learning is that we do not need to change the parameters of pre-trained models for adding new classes to the system. 
Besides, compared with the fine-tuning method, our meta-learning-only method produces better results under the most challenging MSCOCO 1-shot setting, and achieves comparable results under MSCOCO 2/3/5-shot settings, where fine-tuning is prone to overfitting with the small training data. 
Compared with previous meta-learning-only methods~\cite{fan2020few,han2021meta,Han_2021_ICCV}, our method achieves higher results in most of the shots and metrics, especially under the extreme few-shot settings, \eg, 1/2/3-shot on the two benchmarks, where the class semantic information contributes largely to the performance.

With fine-tuning, the performance of our method could be further improved. Our model is better than the strong baseline~\cite{han2021meta}, especially for 1/2/3-shot.

We also achieve much higher performance compared with another multi-modal FSOD method SRR-FSD~\cite{Zhu_2021_CVPR}. Using meta-learning-only, we achieve comparable results, compared with the fine-tuning-based method~\cite{Zhu_2021_CVPR}. With further fine-tuning, our method achieves much higher performance, especially for large shots.

We also provide the visualization of detection results and the failure case analysis in Fig.~\ref{figure_8}.

\begin{table*}[t]
\centering
\footnotesize
\caption{{Few-shot object detection performance (AP50) on the PASCAL VOC dataset, compared with a strong fine-tuning-based model DeFRCN \cite{Qiao_2021_ICCV}.} We report both meta-testing and fine-tuning results. 
}
\addtolength{\tabcolsep}{-0.1cm}
\adjustbox{width=\linewidth}{
\begin{tabular}{c|l|c|ccccc|ccccc|ccccc}
\toprule
& \multirow{2}{*}{Method / Shot} & \multirow{2}{*}{Venue} & \multicolumn{5}{c|}{Novel Set 1} & \multicolumn{5}{c|}{Novel Set 2} & \multicolumn{5}{c}{Novel Set 3} \\ 
& &  &1     & 2     & 3    & 5    & 10   & 1     & 2     & 3    & 5    & 10   & 1     & 2     & 3    & 5    & 10   \\ \midrule
\multicolumn{18}{c}{\textbf{Meta-training the model on base classes, and meta-testing on novel classes}} \\ \midrule
(a) & MM-FSOD & This work & {42.5} & {41.2} & {41.6} & {48.0} & {53.4}  & {30.5} & {34.0} & {39.3} & 36.8 & {37.6} & {39.9} & {37.0} & {38.2} & {42.5} & {45.6} \\ \midrule
\multicolumn{18}{c}{\textbf{Fine-tuning the model on novel classes, and testing on novel classes}} \\ \midrule
(b) & DeFRCN w/o PCB \cite{Qiao_2021_ICCV} & ICCV 2021 & 56.4 & 55.9 & 61.2 & 66.0 & 67.1   & 34.6 & 45.3 & 50.4 & 53.0 & 53.2    & 50.7 & 51.7 & 56.8 & 60.2 & 62.0\\
(c) & DeFRCN \cite{Qiao_2021_ICCV} & ICCV 2021 & 59.0 & 58.6 & 63.7 & 68.0 & 67.3 & 35.5 & 45.1 & 50.9 & 54.5 & 54.6 & 53.4 & 53.6 & 56.5 & 60.1 & 61.9\\
(d) & MM-FSOD & This work & 46.8 & 55.2 & 61.3 & 65.8 & 66.0    & 31.2 & 37.3 & 46.9 & 49.2 & 51.2    & 41.3 & 47.2  & 53.8 & 59.6 & 59.3 \\ 
(e) & MM-FSOD + DeFRCN & This work & \textbf{59.4} & \textbf{59.5} & \textbf{64.6} & \textbf{68.7} & \textbf{68.4} & \textbf{36.0} & \textbf{45.5} & \textbf{51.5} & \textbf{55.0} & \textbf{55.2} & \textbf{54.2} & \textbf{53.7} & \textbf{57.5} & \textbf{60.8} & \textbf{62.5} \\
\bottomrule
\end{tabular}}
\label{tab:main_voc_suppl}
\end{table*}

\begin{table*}[t]
\centering
\footnotesize
\addtolength{\tabcolsep}{-0.15cm}
\caption{Few-shot object detection performance on the MSCOCO dataset, compared with a strong fine-tuning-based model DeFRCN \cite{Qiao_2021_ICCV}. We report both meta-testing and fine-tuning results.
}
\adjustbox{width=\linewidth}{
\begin{tabular}{c|l|ccc|ccc|ccc|ccc|ccc|ccc}
\toprule
& &\multicolumn{3}{c|}{1-shot} & \multicolumn{3}{c|}{2-shot} & \multicolumn{3}{c|}{3-shot} &\multicolumn{3}{c|}{5-shot} & \multicolumn{3}{c|}{10-shot} & \multicolumn{3}{c}{30-shot} \\
& Method & AP & AP50 & AP75 & AP & AP50 & AP75 & AP & AP50 & AP75 & AP & AP50 & AP75 & AP & AP50 & AP75 & AP & AP50 & AP75 \\ \midrule
\multicolumn{20}{c}{\textbf{Meta-training the model on base classes, and meta-testing on novel classes}} \\ \midrule
(a) & MM-FSOD & {5.6} & {11.0} & {5.2}   & {7.9} & 15.3 & {7.4}  & {9.4} & {18.3} & {8.9}   & {10.5} & {19.7} & {9.5}  & {10.8} & {20.5} & {10.2}   & {12.3} & 22.8 & {11.8} \\ 
\midrule
\multicolumn{20}{c}{\textbf{Fine-tuning the model on novel classes, and testing on novel classes}} \\ \midrule
(b) & DeFRCN w/o PCB \cite{Qiao_2021_ICCV} & 5.4 & 9.4 & 5.6 & 9.3 & 16.6 & 9.3 & 12.2 & 22.0 & 12.3 & 14.6 & 27.0 & 14.3 & 17.5 & 32.2 & 16.8 & 21.4 & 38.2 & 21.1 \\
(c) & DeFRCN \cite{Qiao_2021_ICCV} & 6.3 & 11.2 & 6.5 & 10.9 & 19.7 & 10.7 & 13.4 & 24.3 & 13.4 & 15.7 & 29.4 & 14.8 & 18.2 & 34.2 & 16.8 & 22.1 & 39.5 & 21.7 \\
(d) & MM-FSOD & {5.4} & {11.3} & 4.8    & {8.4} & {17.2} & {7.0}   & {10.5} & {21.4} & {8.6}   & {11.4} & {23.8} & {9.3}    & {13.3} & {27.5} & {11.2}   & {17.2} & {33.3} & {16.0} \\ 
(e) & MM-FSOD + DeFRCN & \textbf{6.5} & \textbf{11.5} & \textbf{6.6} & \textbf{11.1} & \textbf{19.9} & \textbf{10.9} & \textbf{13.6} & \textbf{24.5} & \textbf{13.6} & \textbf{16.1} & \textbf{29.7} & \textbf{15.6} & \textbf{18.7} & \textbf{34.6} & \textbf{17.7} & \textbf{22.5} & \textbf{40.1} & \textbf{22.2} \\
\bottomrule
\end{tabular}}
\label{tab:main_coco_suppl}
\end{table*}

\textbf{Comparison with DeFRCN \cite{Qiao_2021_ICCV}.}
As far as we know, DeFRCN \cite{Qiao_2021_ICCV} reports the best fine-tuning results on the two FSOD benchmarks. DeFRCN is built upon a simple fine-tuning baseline model TFA \cite{wang2020few}, by first learning the traditional object detection model on the data-abundant base classes, and then fine-tuning the model on the few-shot novel classes. The contributions of DeFRCN come from two parts, the Gradient Decoupled Layer (GDL) and the Prototypical Calibration Block (PCB). (1) The GDL adjusts the degree of decoupling of the backbone, RPN, and R-CNN through gradient. In practice, stop-gradient is performed between RPN and backbone, and scale-gradient is performed between RCNN and backbone. Moreover, during few-shot fine-tuning, the backbone feature extractor is fine-tuned and the RoI feature extractor is fixed. The dropout layer is also used before the final multi-class classifier. All these techniques contribute to the final strong performance. (2) The PCB introduces a post-processing score calibration model by fusing the fine-tuning-based single-branch model with a two-branch metric-learning-based model, using the ImageNet pre-trained weight.

We provide our full results on the two FSOD benchmarks in Table~\ref{tab:main_voc_suppl} and \ref{tab:main_coco_suppl}, compared with DeFRCN \cite{Qiao_2021_ICCV}. The major findings are,

(1) The highlight of our work is to combine meta-learning with prompt-based learning for multi-modal FSOD without fine-tuning. Our model can easily include novel classes during meta-testing without tuning the parameters of the pre-trained models. However, DeFRCN needs fine-tuning to detect novel classes, which usually requires large computational resources for model training.

(2) Our meta-learning-only model achieves better results on the most challenging MSCOCO 1-shot setting, compared with DeFRCN w/o PCB. This result indicates the strong generalization ability of our meta-learning-only model. Although DeFRCN \cite{Qiao_2021_ICCV} introduces novel techniques to better transfer the pre-trained models to few-shot novel classes, the fine-tuning-based methods are still prone to overfitting to the extremely few-shot setting, e.g., 1-shot.

(3) As shown in Table~\ref{tab:main_voc_suppl} (b-c) and Table~\ref{tab:main_coco_suppl} (b-c), the PCB can bring additional improvements for most of the shots and metrics due to the model fusion. Our proposed model belongs to the two-branch metric-learning-based methods, which is complementary to the strong fine-tuning-based single branch model DeFRCN w/o PCB. Therefore, we combine our proposed method with DeFRCN using PCB. We also find that the ImageNet pre-trained model used in the original DeFRCN PCB module is useful. We thus combine the model (c) with (d) for model (e) in both Table~\ref{tab:main_voc_suppl} and \ref{tab:main_coco_suppl}. The final fused models consistently outperform any of the single model (including the original DeFRCN models and our MM-FSOD models) for most of the shots and metrics in the two FSOD benchmarks.

\section{Conclusion} 

We studied multi-modal FSOD, using both few-shot visual examples and class semantic information for detection. Our approach is motivated by the high-level conceptual similarity of meta-learning and prompt-based learning to learn generalizable few-shot and zero-shot object detection models respectively without fine-tuning.
Specifically, we combine the few-shot visual classifier and text classifier learned via meta-learning and prompt-based learning respectively for the multi-modal classifier and detection models. 
Moreover, the meta-learning-based cross-modal prompting is used to generate soft prompts for novel classes present in few-shot visual examples. Knowledge distillation is introduced to learn the prompt generator without using human prior knowledge like class names. Extensive ablations on the two widely used FSOD benchmarks (PASCAL VOC \& MSCOCO) verify the effectiveness of our approach. 

In the future, we would like to extend our work by using other meta-data to assist in detection, \eg, attributes.

\section*{Acknowledgements}
{
\noindent This material is based on research sponsored by Air Force Research Laboratory (AFRL) under agreement number FA8750-19-1-1000. 
The U.S. Government is authorized to reproduce and distribute reprints for Government purposes notwithstanding any copyright notation therein. 
The views and conclusions contained herein are those of the authors and should not be interpreted as necessarily representing the official policies or endorsements, either expressed or implied, of Air Force Laboratory, DARPA or the U.S. Government.
}

\section*{Data Availibility Statement}
The MSCOCO dataset is available at \url{https://cocodataset.org/}, and the PASCAL VOC dataset is available at \url{http://host.robots.ox.ac.uk/pascal/VOC/}.

\bibliographystyle{spbasic}      
\bibliography{MM-FSOD}



\end{document}